# A Hybrid Chimp Optimization Algorithm and Generalized Normal Distribution Algorithm with Opposition-Based Learning Strategy for Solving Data Clustering Problems

Sayed Pedram Haeri Boroujeni[a*], Elnaz Pashaei[b]

[a*]School of Computing, Clemson University, Clemson, SC, USA

Email: shaerib@g.clemson.edu

[b]Department of Software Engineering, Istanbul Aydin University, Istanbul, Turkey

Email: elnazpashaei@aydin.edu.tr

**Abstract:** This paper is concerned with data clustering to separate clusters based on the connectivity principle for categorizing similar and dissimilar data into different groups. Although classical clustering algorithms such as K-means are efficient techniques, they often trap in local optima and have a slow convergence rate in solving high-dimensional problems. To address these issues, many successful meta-heuristic optimization algorithms and intelligence-based methods have been introduced to attain the optimal solution in a reasonable time. They are designed to escape from a local optimum problem by allowing flexible movements or random behaviors. In this study, we attempt to conceptualize a powerful approach using the three main components: Chimp Optimization Algorithm (ChOA), Generalized Normal Distribution Algorithm (GNDA), and Opposition-Based Learning (OBL) method. Firstly, two versions of ChOA with two different independent groups' strategies and seven chaotic maps, entitled ChOA(I) and ChOA(II), are presented to achieve the best possible result for data clustering purposes. Secondly, a novel combination of ChOA and GNDA algorithms with the OBL strategy is devised to solve the major shortcomings of the original algorithms. Lastly, the proposed ChOAGNDA method is a Selective Opposition (SO) algorithm based on ChOA and GNDA, which can be used to tackle large and complex real-world optimization problems, particularly data clustering applications. In this study, eight benchmark datasets including five datasets of the UCI machine learning repository and three challenging shape datasets are used to investigate the general performance of the proposed method. The results are evaluated against seven popular meta-heuristic optimization algorithms and eight recent state-of-the-art clustering techniques. Experimental results illustrate that the proposed work significantly outperforms other existing methods in terms of the achievement in minimizing the Sum of Intra-Cluster Distances (SICD), obtaining the lowest Error Rate (ER), accelerating the convergence speed, and finding the optimal cluster centers.

## 1. Introduction

Over the past few decades, an enormous amount of unstructured information has been widely produced in various fields of science. Due to the making effective use of this information, different methods are necessitated to classify or categorize them. Data clustering is one of the most useful and powerful techniques which has a significant effect on data analysis performance. The application of clustering can be considered in the fields of bioinformatics [1], image analysis [2], signal processing [3], text mining [4], and medicine [5]. Within the field of machine learning, there are two main types of tasks: supervised learning, and unsupervised learning. The main difference between the two approaches is that supervised learning uses labeled data to predict the output values, while the other does not. One of the most common techniques in unsupervised learning is clustering which does not require any prior information about data [6]. It is a process of categorizing a set of data into different clusters, where the data points within a specific cluster must be extremely similar to each other and the data within different clusters must be highly dissimilar to each other.

Generally, there are five various kinds of clustering algorithms including partitional, hierarchical, fuzzy, density-based, and model-based methods. Partitioning algorithm [7] is a type of clustering technique that divides the data points into different groups based on their similarities, where the number of clusters should be pre-specified. In this category, K-means is regarded as the most successful and powerful algorithm. Hierarchical algorithm [8] is an alternative clustering approach that does not need to determine the number of groups for identifying clusters in the dataset. Single-link and complete-link are the two most popular hierarchical algorithms. Fuzzy algorithm [9] is also known as the soft clustering method, whereby, each data point has a probability of belonging to one of the groups. In other words, in soft clustering, each item can belong to more than one cluster, while in hard clustering, each item can only belong to one cluster. The Fuzzy C-Means (FCM) can be considered as the most well-known algorithm of the mentioned category. The density-based algorithm [10] is another type of clustering method that relies on separating the regions with high point density from the regions with low point density. DBSCAN (Density-Based Spatial Clustering and Application with Noise) is an example of a density-based algorithm that is suitable for identifying the clusters of different shapes and sizes in noisy datasets. The last method is model-based clustering algorithm [11] which assumes that the data were generated by a specific model. To this aim, it attempts to find the best fit model and assigns each data to the corresponding cluster.

Despite the fact that traditional clustering algorithms are an effective strategy in a wide range of applications, they are faced with some issues and limitations. The first and foremost disadvantage is that they are completely dependent on the initialization parameters. The second drawback is that they suffer from the local optima problem. Moreover, they are not appropriate for large datasets due to their high computational cost and complexity. Lastly, there is not exist an efficient method for partitioning the set of data into different clusters when the dataset contains overlapping areas. To overcome the previous drawbacks, the clustering problem can be assigned as an optimization problem that the goal is to maximize or minimize the objective function. If the objective function is well-designed and capable of capturing the significant features from the datasets, then an efficient clustering technique can be expected. Within the field of data clustering, objective functions are commonly defined based on the similarity measure (distance metrics). The aim of the objective function is to identify the best possible cluster center, where the data points within a cluster should be closest to its centroid. In addition to the importance of the objective function, the optimization procedure can be modified to observe a significant impact on the clustering performance. The key component of this modification is Metaheuristic Optimization Algorithms (MOAs), which are employed to discover the search space for exploiting the best possible solution. MOAs applications are widely used in various fields of science, such as an ant-inspired algorithm for the collective discovery of workflow services [12], a Volcano Eruption Algorithm (VEA) for solving optimization problems [13], Ant-based Replication and Mapping protocol (ARMAP) for classification or clustering of Grid resources (Forestiero et al., 2005) and Horse Herd Optimization Algorithm (HHOA) for medical solving problems [14]. So, data clustering is a challenging task where the MOAs play a significant role in solving the specified problem. Through the analysis of recent scientific literature, it can be observed that MOAs have a number of important characteristics, particularly when data clustering is considered as an optimization problem. Totally, details are organized below:

1) MOAs maintain the simple concept and structure.

2) MOAs have the ability to escape from a local optima problem.

3) MOAs are a type of derivation-free algorithm that is useful for solving real-world problems.

4) MOAs can be applied to a wide range of problems without requiring any structural changes due to their flexibility and adaptability.

In general, MOAs are categorized into four major groups including human-inspired algorithms, physics-inspired algorithms, Evolutionary Algorithms (EAs), and Swarm Intelligence Algorithms (SIAs). The first group is human-inspired optimization algorithms that mimic humans' behaviors in their social activities. The most recent and well-known algorithms of this group are Poor and Rich Optimization (PRO) [15], Human Mental Search (HMS) [16], Student Psychology-Based Optimization (SPBO) [17], and Algorithm of the Innovative Gunner (AIG) [18]. The next group is physics-inspired optimization algorithms where the physical phenomena are the primary source of their inspiration. Henry Gas Solubility Optimization (HGSO) [19], Equilibrium Optimizer (EO) [20], Archimedes Optimization Algorithm (AOA) [21], and Thermal Exchange Optimization (TEO) [22] are a few examples of physics-based algorithms. The third group is EAs, also known as bio-inspired algorithms, which are inspired by the ideas of biological evolution. It can be referred to as Genetic Algorithm (GA) [23], Porcellio Scaber Algorithm (PSA) [24], Sin Cosine Algorithm (SCA) [25], and Degree-Descending Search Evolution (DDSE) [26] among the conventional algorithms of this group. SIAs are the last group of MOAs, imitating the collective (swarm) behavior of different entities, especially those species who rely on consensus decision-making in their processes. Swarm behavior in birds, fishes, tetrapods, and insects is called flocking, schooling, herding, and colonies, respectively. Some of the common algorithms in this category include Particle Swarm Optimization (PSO) [27], Harris Hawks Optimization (HHO) [28], Chimp Optimization Algorithm (ChOA) [29] Binary Dragonfly Algorithm (BDA) [30], Butterfly Optimization Algorithm (BOA) [31], and Sparrow Search Algorithm (SSA) [32].

Eventually, the successful application of MOAs in dealing with various optimization cases motivated us to incorporate SIAs into data clustering problems. Based on this idea, we utilized the recently proposed SIA called Chimp Optimization Algorithm (ChOA) to solve data clustering problems [33]. The major advantages of using SIAs over other MOAs in data clustering tasks can be summarized as follows: (i) simple structure, which enables scientists to implement SIAs more easily; (ii) few numbers of control parameters: this reduces the high computational complexity of SIAs associated with data clustering algorithms; (iii) limited critical operators: SIAs require less computational burden for finding the best solution compared to other existing EAs such as GA (crossover, mutation, etc); (iv) memory-based mechanisms, which refers to the SIA's ability to memorize the useful information during each iteration. Despite the significant achievements of ChOA in dealing with data clustering problems, the fundamental deficiencies of MOAs such as premature convergence, trapping in local optima, inability to maintain a promising balance between local search and global search, and not always converging to the best global solution motivates us to develop a hybrid approach to mitigate the limitations of individual algorithms. Moreover, the No-Free Lunch (NFL) theory [34] states that no single MOAs can efficiently solve the extensive range of complicated optimization problems. Consequently, this paper introduces a novel hybrid clustering method, called ChOAGNDA by the combination of ChOA and GNDA algorithms with Opposition-Based Learning (OBL) strategy for dealing with more complex clustering problems.

The main objective of this study is to develop a reliable clustering algorithm that can be able to categorize a set of data into different clusters accurately. The proposed algorithm not only improved the clustering performance but also reduced the number of misclassified data. Meanwhile, it can successfully deal with the local minimum problem and slow convergence speed. Furthermore, the proposed algorithm can be applied to a wide range of clustering problems including challenging shape datasets and high-dimensional benchmark datasets. In brief, ChOAGNDA is designed to transform clustering into an optimization problem, and then optimize the specified objective function to find the appropriate cluster centers. Following that, all of the data in a dataset are divided into different clusters based on the principle of minimum distances between each data and its centroid. To the best of our knowledge, this is the first time that ChOA, GNDA, and OBL technique have been combined and formed a hybrid algorithm to solve data clustering problems. The major contributions of this paper are as follows:

- This study suggested a new hybrid approach based on ChOA and GNDA algorithms with a selective OBL strategy (ChOAGNDA) for solving complex data clustering problems.

- The combination of ChOA and GNDA strategies enhanced the algorithm's performance in achieving the best possible results in terms of convergence behaviors, search efficiency, and evaluation measures (SICD and ER). Moreover, the use of OBL technique allowed for a stable balance between exploration and exploitation phases to avoid local optima.

- Two modifications are incorporated into the ChOA algorithm to improve its preformance in dealing with challenging clustering cases.

- i. Firstly, we proposed another version of ChOA called ChOA(II) and compared it to the proposed ChOA(I) in Ref. [33]. These two ChOA versions are completely different in terms of global and local search, which ensures the best clustering performance.
- ii. Secondly, we employed seven chaotic maps with different characteristics in ChOA to guarantee the best possible result. ChOA is extremely sensitive to the chaotic value, so this is a useful idea to improve its capability during the optimization process.

- Finally, the performance of the proposed method is evaluated using five benchmark and three shape datasets, and the results are compared against several optimization algorithms and well-regarded hybrid approaches.

The remainder of this study is outlined as follows: Section 2 reviews the related literature. Section 3 presents preliminaries including data clustering process, Chimp Optimization Algorithm (ChOA), Generalized Normal Distribution Algorithm (GNDA), and Opposition-based learning (OBL) mechanism. In section 4, the details of the clustering problem and proposed methods are presented. The experimental results and statistical analysis are elaborated in section 5, while the conclusion and future directions are summarized in section 6.

**2. Review of the Related Literature**

During the last few years, many studies focused on data clustering not only as an important task of data mining but also as a dynamic way for testing optimization algorithms' efficiency. This is a prominent area that has recently attracted a lot of attention from researchers and professionals from various fields of science. In order to provide a clear perspective of clustering methods, Table (1) categorizes and lists the most significant researches that have been proposed to solve data clustering problems.

In the available literature, the diversity of optimization algorithms for data clustering problems is numerous. Class Topper Optimization (CTO) [35] is a kind of human-based optimization algorithms, which is inspired by the students' learning intelligence. Students in the same classroom make effort to increase their knowledge in order to succeed in their exams. Besides, various examinations are performed to assess the students' performance to identify the best student in each class. The best student of the class is referred to as a "class topper", and the CTO algorithm aims to enhance the class topper's performance during all the further examinations process. Das et al. 2020 [35] considered the CTO algorithm as an optimization method to solve clustering problems. The proposed clustering algorithm requires the encoding of four essential parameters: students, class topper, courses, and exams. Students refer to the search agents which are assigned to find the optimal solution. Class topper refers to the best cluster center that has the minimum sum of intra-cluster distances. The number of courses taken by each student indicates the number of clusters, and the number of examinations shows the number of iterations. The proposed algorithm is evaluated regarding the SICD result, Average Percentage of Error (APE), and convergence rate. The main advantage of the CTO algorithm is the ability to find the best cluster centers while maintaining the minimum range of error rates. However, the main drawback of this algorithm is its inability to deal with non-spherical data.

**Table 1.** Previous literature of data clustering using Meta-heuristic optimization algorithms.

| Category | Algorithm (s) Name | Arbitrary Name | First Author | Year | Ref. |
|---|---|---|---|---|---|
| Human-based | Stem Cells Algorithm | SCA | Taherdangkoo M | 2012 | [36] |
| | Heart Optimization Algorithm | HOA | Hatamlou A | 2014 | [37] |
| | Imperialist Competitive Algorithm | ICA | Zadeh MR | 2014 | [38] |
| | Class Topper Optimization | CTO | Das P | 2018 | [35] |
| | Teaching Learning Based Optimization | TLBO | Naik A | 2020 | [39] |
| Physics-based | Big Bang–Big Crunch | BBBC | Hatamlou A | 2011 | [40] |
| | Gravitational Search Algorithm | GSA | Hatamlou A | 2011 | [41] |
| | Black Hole | BH | Hatamlou A | 2013 | [42] |
| | Multi-Verse Optimizer | MVO | Shukri S | 2018 | [43] |
| | Improved Black Hole | IBH | Deeb H | 2021 | [44] |
| Evolutionary | Genetic Algorithm | GA | Murty MN | 2008 | [45] |
| | Quantum Evolutionary Algorithm | QEA | Ramdane C | 2010 | [46] |
| | Biogeography-Based Optimization | BBO | Hammouri AI | 2014 | [47] |
| | Sine Cosine Algorithm | SCA | Kumar V | 2017 | [48] |
| | Gradient Evolution | GE | Kuo RJ | 2020 | [49] |
| Swarm-based | Particle Swarm Optimization | PSO | Van DW | 2003 | [50] |
| | Cuckoo Search Algorithm | CSA | Manikandan P | 2014 | [51] |
| | Grasshopper Optimization Algorithm | GOA | Lukasik S | 2017 | [52] |
| | Moth–Flame Optimization | MFO | Shehab M | 2020 | [53] |
| | Modified Grey Wolf Optimizer | MGWO | Ahmadi R | 2021 | [54] |
| Hybrid | Genetic Algorithm, Particle Swarm Optimization | GAPSO | Kuo RJ | 2010 | [55] |
| | Krill Herd Algorithm, Hybrid Function | MMKH | Abualigah LM | 2018 | [56] |
| | K-means, Ant Lion Optimization | K-ALO | Majhi SK | 2018 | [57] |
| | Firefly Algorithm, Particle Swarm Optimization | FAPSO | Agbaje MB | 2019 | [58] |
| | Whale Optimization Algorithm, Tabu Search | WOATS | Ghany KK | 2020 | [59] |

Multi-Verse Optimizer (MVO) [60] is a type of physics-based optimization algorithm that is based on the theory of multi-verse in astrophysics. According to this theory, three kinds of holes including white holes, black holes, and worm holes are used to create an interaction between different universes. In more particular, various universes are formed through different big bangs, whereby, each universe has the possibility of having white holes, black holes, and worm holes. The responsibility of each component is as follows: white holes emit the objects, black holes absorb the objects, and worm holes prepare a connection between the universes by making a tunnel through space-time. The algorithm's objective is to find the best universe and then improve the performance of the corresponded universe during the interaction process. Shukri et al. 2018 [43] employed the MVO algorithm as an optimization method to tackle data clustering problems in two distinct approaches. The first approach is a type of static clustering method (SCMVO), in which the number of clusters requires to be predefined in advance. In contrast, the second approach is a type of dynamic clustering method (DCMVO) that the number of clusters is determined by the algorithm. The proposed clustering algorithm needs encoding of five major parameters: universes, white holes, black holes, time, and inflation rate. Universes refer to the clustering solutions, where each universe is regarded as the center of candidate solutions. White holes represent the best set of cluster centers that has the highest objective value, while black holes show the worst set of cluster centers with the lowest objective value. Time and inflation rate indicate the number of iterations and the objective value of each universe, respectively. Finally, the performance of the MVO algorithm is evaluated in terms of purity (the percentage of correctly classified objects), entropy (the semantic distribution of the objects within each cluster), and the convergence rate. The results show that the proposed algorithm obtains an acceptable value of purity and entropy in only half of the given datasets, besides it has the slowest convergence rate among all the algorithms.

Grey Wolf Optimizer (GWO) [61] is a kind of SIAs, which simulates the hunting behavior of grey wolves. There are four kinds of grey wolves in the leadership hierarchy entitled alpha, beta, delta, and omega. Alpha is the highest-ranked wolf (leader) in the pack, and he/she is responsible for managing the members of the pack. The second-ranked wolf is beta that assists the alpha wolf in different duties such as decision-making, tracking, hunting, etc. Omega is the lowest ranking wolf in the hierarchy of grey wolves that plays an important role in maintaining the pack's survival. Lastly, any wolf that does not belong to one of the previous categories is referred to as delta, which is ranked third among the hierarchy of grey wolves. The process of grey wolf hunting is divided into three major phases: searching, surrounding, and attacking. The purpose of the GWO algorithm is to find the prey's location and finish the hunting process by attacking the prey. Ahmadi et al. 2021 [54] proposed Modified Grey Wolf Optimizer (MGWO) as an optimization method for solving data clustering limitations. In this algorithm, a new control parameter is utilized to make an appropriate balance between the exploration and exploitation parts. It is an effective way to improve the GWO performance. In the proposed clustering algorithm, different types of grey wolves are considered as the center of candidate solutions (cluster centers). Alpha wolf indicates the best solution, beta wolf refers to the second-best solution, delta wolf is the third-best solution, and the rest of the solutions belong to omega wolves. The algorithm's objective is to improve the performance of the alpha wolves in each iteration. Eventually, the efficiency of the proposed clustering algorithm is evaluated regarding the objective value (sum of intra-cluster distances) and the error rate (the percentage of misclassified objects). The conducted experiments illustrate that MGWO achieves promising results in most of the given datasets.

Firefly Algorithm (FA) [62] and Particle Swarm Optimization (PSO) algorithm are the two well-known SIAs that have been successfully employed to solve a wide range of optimization problems. FA mimics the luminous behavior of fireflies in nature, while PSO imitates the social behavior of swarms such as bird flocks, schooling fish, etc. The basic principles of FA are as follows: first of all, fireflies are a type of unisex species that their mating strategy is based on the quantity of their light intensity. Secondly, the brightest firefly moves randomly, while the low brightness fireflies move towards the high brightness fireflies. Lastly, the light intensity of each firefly is specified by its distance from the other fireflies and it is considered as the objective function. If the distance is increased, the light intensity will be decreased. On the other hand, the main principle of the PSO algorithm is associated with two properties namely velocity and position. The PSO algorithm starts with an initial random population of candidate solutions, where each one of them has the potential of being an optimal solution. Each candidate solution is referred to as a particle. Particles differ in terms of velocity and they move through the search space according to their velocity values. In other words, the velocity directs the movement of each particle to its optimum position by following two parameters. The first parameter is called local best (Pbest) which represents the best position of each particle, and the second parameter is called global best (Gbest) which refers to the best position of neighbor particles. Finally, the position of each particle is determined by its velocity. Agbaje et al. 2019 [58] designed an efficient hybrid algorithm called FAPSO, in which a modified FA is combined with PSO to handle the clustering problems. FAPSO algorithm employs the FA as the primary search algorithm at the early stages of the search process and then continues with PSO to discover the best optimal solutions (cluster centers). Experimental results prove the efficiency of FAPSO algorithm regarding the convergence speed and clustering quality, compared to the other clustering algorithms such as FA and PSO. It should be noted that the significant disadvantage of FAPSO algorithm is that the combination of two MOAs incredibly increases the amount of computational complexity.

In summary, the main highlights of the technical limitations associated with existing approaches are listed below. This motivates us to propose a new clustering algorithm which can overcome the shortcomings of individual techniques.

- The majority of existing techniques still suffer from trapping in local minima.
- They have a slow convergence rate especially in solving high-dimensional problems.
- They have a high execution time, which makes them unable to find optimal solutions in a reasonable amount of time.
- Due to the number of control parameters, some clustering strategies are more computationally expensive than other approaches.
- Among existing algorithms, many of them fail to minimize the SICD value while maintaining the ER range as small as possible.
- Most of the proposed techniques are faced with some deficiencies, such as dealing with various types of data (shape, spherical, non-spherical, etc).

## 3. Preliminaries

This section describes the prerequisite theories and algorithms in the following subsections. Section 3.1: data clustering background, Section 3.2: chimp optimization algorithm, Section 3.3: generalized normal distribution algorithm, and Section 3.4: opposition-based learning mechanism.

### *3.1. Data Clustering Analysis*

The main objective of data clustering is to partition N data objects into K clusters. N can be defined as $N = \{D_i : D_1, D_2, \ldots, D_N\}$, where N indicates the number of samples and $D_i$ represents the position of each sample. Furthermore, K can be specified as $K = \{C_j : C_1, C_2, \ldots, C_K\}$, where K shows the number of clusters and $C_j$ is the position of each cluster's center. Data points within the same cluster should be highly similar to each other, while data points within different clusters should be dissimilar as much as possible. To this aim, various distance metrics such as Euclidean distance [63], Manhattan distance [64], and Minkowski distance [65] are proposed to determine the similarity measure. In this work, we utilize the Euclidean distance metric to calculate the distances between m-dimensional data objects and their cluster centers, as follows:

$$\text{dis}(D_i, C_j) = \sqrt{\sum_{n=1}^{m}(D_{in} - C_{jn})^2} \tag{1}$$

To achieve the highest performance in data clustering, the similarity of data objects within the same cluster should be maximized and the similarity of data objects within different clusters needs to be minimized. For this purpose, Sum of the Intra-Cluster Distances (SICD) based on Euclidean distance is considered as the objective function, and the target is to minimize the SICD. The objective function of our proposed clustering algorithm is defined according to Equation (2).

$$f(D,C) = \sum_{i=1}^{N}\sum_{j=1}^{K} W_{ij}\|D_i - C_j\|^2 \tag{2}$$

Where $f(D, C)$ represents the objective value which is also known as cluster integrity, $\|D_i\text{-}C_j\|$ indicates the Euclidean distance between a data sample and the cluster center, and $W_{ij}$ is the association weight that could be one (if sample i is assigned to cluster j) or zero (if sample i is not assigned to cluster j).

*3.2. Chimp Optimization Algorithm*

Chimp Optimization Algorithm (ChOA) is one of the latest SIAs, introduced by Khishe and Mosavi in 2020 [29]. This algorithm imitates the social behavior of chimps including their individual intelligence and sexual motivation while they are in a hunting group. Each group contains various types of chimps that are not similar to each other regarding their ability and intelligence. Although they are all carrying out their responsibilities as a member of the group, they have their own strategy to discover the environment. A chimp colony consists of four kinds of chimps including drivers, barriers, chasers, and attackers. They all have unique skills and each skill can be efficient in a specific part of the hunting process. Drivers pursue the prey without trying to overtake it. Barriers construct a dam on top of the trees to prevent the prey from moving forward. Chasers are responsible for following the prey to overtake it. Lastly, attackers attack the prey by closing its escape route, which forces it back to the chasers or leads it into the trap. The attackers' performance is closely related to their age, intelligence, and physical strength. In general, the chimps' hunting process consists of two critical parts namely exploration and exploitation. The exploration part refers to driving, chasing, and blocking the prey, while the exploitation part refers to attacking and hunting the prey. The significant point in achieving the high performance of ChOA is that the appropriate balance should be chosen between exploration and exploitation.

Preparing a mathematical model for the ChOA algorithm needs five independent parts which are formulated as follows:

### 3.2.1. Encircling Part

As noted previously, the prey is encircled by the chimps during the hunting process. Equations (3) and (4) represent the mathematical model of the driving and chasing mechanism:

$$d = |c.X_{prey}(t) - M.X_{chimp}(t)| \qquad (3)$$

$$X_{chimp}(t+1) = X_{prey}(t) - a.d \qquad (4)$$

Where t, $X_{prey}$, and $X_{chimp}$ show the number of current iterations, the position vector of prey, and the vector of chimp position, respectively. a, c, and M parameters are the coefficient vectors and they are computed using Equations (5)-(7), respectively.

$$a = 2.f.r_1 - f \qquad (5)$$

$$c = 2.r_2 \qquad (6)$$

$$M = \text{Chaotic\_Value} \qquad (7)$$

Where f indicates the boundary range of non-linearly that is declined from 2.5 to 0 over the course of iterations. $r_1$ and $r_2$ are the random vectors between 0 to 1, and M is a chaotic vector that demonstrates the impact of chimp sexual behavior on the hunting process. It should be stated that each chimp has multiple options of selecting any possible positions around the prey's location. To this aim, selecting or changing the position of each chimp is adjusted according to the $r_1$ and $r_2$ vectors, and the movement direction of each chimp is determined through the a and c values. Therefore, each chimp is able to randomly update its location based on the prey's position by using Equations (3) and (4).

### 3.2.2. Exploitation Part

It refers to the attacking behavior of chimps in which the entire process is guided by the attackers. In other words, attacker chimps are in charge of the main tasks, while the other chimps (drivers, barriers, and chasers) assist them in the hunting process. As mentioned previously, this is an unsupervised searching where is no information available about the prey's location (optimal solution). Therefore, it is supposed that the best solutions are obtained by the first driver, chaser, barrier, and attacker (search agents) and the rest of the chimps update their positions based on the location of the best chimps. Equations (8)-(10) are the mathematical model of the attacking mechanism:

$$d_{Attacker} = |c_1 X_{Attacker} - M_1 X|$$
$$d_{Barrier} = |c_2 X_{Barrier} - M_2 X|$$ (8)
$$d_{Chaser} = |c_3 X_{Chaser} - M_3 X|$$
$$d_{Driver} = |c_4 X_{Driver} - M_4 X|$$

$$X_1 = X_{Attacker} - a_1(d_{Attacker})$$
$$X_2 = X_{Barrier} - a_2(d_{Barrier})$$ (9)
$$X_3 = X_{Chaser} - a_3(d_{Chaser})$$
$$X_4 = X_{Driver} - a_4(d_{Driver})$$

$$X(t+1) = \frac{X_1 + X_2 + X_3 + X_4}{4}$$ (10)

Where parameters a, c, and M represent the coefficient vectors which were calculated by Equations (4)-(6), X refers to the chimp positions, and d is the distance between each chimp and its prey. To clarify the process of updating positions in ChOA, Figure (1) is provided to depict how a chimp's position is updated based on the location of other chimps. It is observed that the prey's location is approximated by the four of the best chimps including the driver, chaser, barrier, and attacker, then the rest of the chimps update their positions through the random area surrounding the prey.

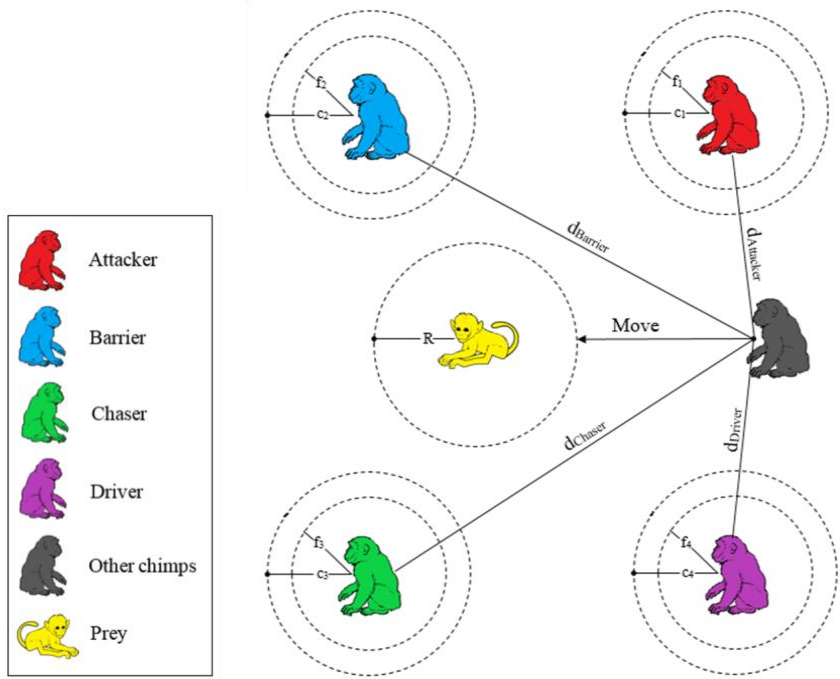

**Figure 1.** Position updating in Chimp optimization algorithm.

### 3.2.3. Utilization Part

It refers to the last stage of the hunting process where the chimps attack the prey to terminate the hunting event by obtaining the meat. In the mathematical models for formulating the attacking part, a is a random number between the range of [-2f, 2f] and f is a boundary range of [0, 2.5] which is declined during the iterations. Whenever the |a| value is in the interval of [-1, 1], it indicates that the chimp's next location could be anywhere between its current position and the prey's position. To this aim, the inequality |a|<1 obliges the chimps to move forward the prey, while inequality |a|>1 obliges the chimps to move backward in search of another prey. Figure (2) shows the position updating mechanism of chimps based on the value of |a|.

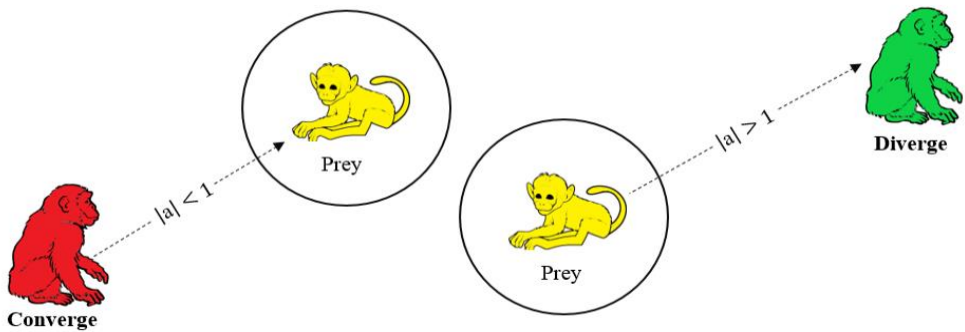

**Figure 2.** Effect of the |a| value on position updating procces.

### 3.2.4. Exploration Part

As previously stated, the ChOA may be trapped in local optima due to its updating mechanism, which is based on the positions of the driver, chaser, barrier, and attacker chimps in the search space. Hence, the algorithm must devote more attention to the exploration part to prevent the mentioned issue. The exploration part refers to the searching task for finding the prey's location in order to finish the hunting process. In the mathematical models, two key parameters can affect the performance of the exploration part. The first parameter is a, which represents a random value that can be greater than 1 or less than -1. The inequality |a|<1 means that the chimps should converge to the prey, meanwhile |a|>1, shows that the chimps should diverge from the prey to find more suitable prey. The second parameter is c, which is a random vector between the range of 0 and 2. The inequalities c >1 and c <1 are able to strengthen and weaken the effect of prey's position on distance calculation, respectively. Additionally, the c parameter requires the generation of random values over the course of iterations, which leads to getting rid of the local minima problem.

### 3.2.5. Social Motivation Part

It refers to the sexual incentive of chimps, in which they attempt to acquire meat chaotically to exchange it for their social needs such as sex and grooming. This chaotic behavior helps them solve two significant issues: slow convergence rate and trap in local minima. Currently, many types of chaotic maps exist that can be applied to the ChOA algorithm. The details of the chaotic maps employed for the proposed clustering algorithm are explained in the next section. The mathematical model of simultaneous updating position is defined in Equation (11). In this model, the possibility of 0.5 is considered for selecting between the normal updating position and the chaotic updating mechanism.

$$X_{Chimp}(t+1) = \begin{cases} X_{prey}(t) - a.d & \text{if } \mu < 0.5 \\ Chaotic\_Value & \text{if } \mu \geq 0.5 \end{cases} \quad (11)$$

Where μ is a random value between 0 and 1. Eventually, Algorithm (1) represents the pseudo-code of ChOA in detail.

---

**Algorithm 1.** Pseudo-code of ChOA

---

1.     Initialize the population $X_i$ (i = 1, 2, …, n)
2.     Initialize **f**, **a**, **c**, and **M**
3.     Calculate the position of each chimp
4.     Divide chimps randomly into different groups until the condition is satisfied
6.     Calculate the fitness of each chimp
7.     $X_{Attacker}$ = the best search agent
8.     $X_{Chaser}$ = the second-best search agent
9.     $X_{Barrier}$ = the third-best search agent
10.    $X_{Driver}$ = the fourth-best search agent
11.       **While** (t < maximum number of iterations)
12.         **for** each chimp:
13.            Extract the chimp's group
14.            Use its group strategy to update **f**, **c**, and **M**
15.            Use **f**, **c**, and **M** to calculate **a** and **d**
16.         **end for**
17.          **for** each search chimp
18.                 **if** (μ < 0.5)
19.                     **if** (|a| < 1)
20.            Update the position of the current search agent by Equation (4)
21.                     **else if** (|a| > 1)
22.                           Select a random search agent
23.                     **end if**
24.                 **else if** (μ > 0.5)
25.            Update the position of the current search agent by Equation (11)
26.                 **end if**
27.           **end for**
28.        Update **f**, **a**, **c**, and **M**
29.        Update $X_{Attacker}$, $X_{Chaser}$, $X_{Barrier}$, and $X_{Driver}$
30.        t = t + 1
31.     **end while**
32.    **return** $X_{Attacker}$

### 3.3. Generalized Normal Distribution Algorithm

The Generalized Normal Distribution Algorithm (GNDA) is another recent MOAs, proposed by Zhang, Jin, and Mirjalili in 2020 [66]. The framework of the original GNDA is divided into four subsections. Section 3.3.1 discusses the motivation of GNDA, Section 3.3.2 explains the local search strategy, Section 3.3.3 provides the explanation of global search strategy, and Section 3.3.4 investigates the screening mechanism of GNDA.

### 3.3.1. Inspiration

The primary source of its inspiration is normal distribution theory, also known as Gaussian distribution, which plays an important role in defining natural phenomena. The mathematical formulation of the normal distribution is modeled according to the Equation (12).

$$f(x) = \frac{1}{\sqrt{2\pi\delta}} \exp\left(-\frac{(X-\mu)^2}{2\delta^2}\right) \tag{12}$$

Where X represents a normal random variable, $\mu$ indicates a probability distribution, and $\delta$ is a scale parameter.

### 3.3.2. Local Exploitation

As previously mentioned, the optimization process of GNDA is divided into two parts: local search strategy and global search strategy. This section explains local exploitation, which is the process of determining optimal solutions in the search space among all individual positions. According to the correlation between the population distribution and normal distribution, the generalized normal distribution model is mathematically formulated as Equation (13):

$$V_i^t = \mu_i + \delta_i \times \eta, \qquad i = 1,2,3,\ldots,N \tag{13}$$

Where $V_i^t$ represents the trial vector of the $i_{th}$ solution at $t_{th}$ iteration, $\mu_i$ indicates the generalized mean value of the $i_{th}$ solution, $\delta_i$ refers to the generalized standard deviation, and $\eta$ is the penalty factor. In addition, the value of $\mu_i$, $\delta_i$, and $\eta$ is obtained from Equations (14) to (16).

$$\mu_i = \frac{1}{3}(X_i^t + X_{Best}^t + M) \tag{14}$$

$$\delta_i = \sqrt{\frac{1}{3}[(X_i^t - \mu)^2 + (X_{Best}^t - \mu)^2 + (M - \mu)^2]} \tag{15}$$

$$\eta = \begin{cases} \sqrt{-\log(\lambda_1)} \times \cos(2\pi\lambda_2), & \text{if } A \leq B \\ \sqrt{-\log(\lambda_1)} \times \cos(2\pi\lambda_2 + \pi), & \text{Otherwise} \end{cases} \tag{16}$$

Where A, B, $\lambda_1$, and $\lambda_2$ reperesnt a random number between the range of [0, 1], $X_{Best}^t$ indicates the best position, and M is the mean value of the current population. Moreover, M is obtained from Equation (17).

$$M = \frac{\sum_{i=1}^{N} X_i^t}{N} \tag{17}$$

### 3.3.3. Global Exploration

This section describes global exploration, which is the process of exploring the search space to discover promising areas. Therefore, the global exploration in GNDA can be mathematically formulated as Equation (18):

$$V_i^t = X_i^t + \beta \times (|\lambda_3| \times V_1) + (1 - \beta) \times (|\lambda_4| \times V_2) \tag{18}$$

Where $\lambda_3$ and $\lambda_4$ represent the random numbers, which are controlled through the normal distribution. β is another random number between the range of [0, 1], and $V_1$ and $V_2$ are trial vectors. Furthermore, $V_1$ and $V_2$ are calculated according to the Equations (19) and (20).

$$V_1 = \begin{cases} X_i^t - X_{P1}^t, & \text{if } f(X_i^t) < f(X_{P1}^t) \\ X_{P1}^t - X_i^t, & \text{Otherwise} \end{cases} \tag{19}$$

$$V_2 = \begin{cases} X_{P2}^t - X_{P3}^t, & \text{if } f(X_{P2}^t) < f(X_{P3}^t) \\ X_{P3}^t - X_{P2}^t, & \text{Otherwise} \end{cases} \tag{20}$$

Where P1, P2, and P3 represent the random integers in the interval of [0, N], which should be P1 ≠ P2 ≠ P3 ≠ i.

### 3.3.4. Screening Mechanism of GNDA

GNDA, like other optimization algorithms, starts by generating a random population by using Equation (21).

$$X_{i,j}^t = L_j + (U_j - L_j) \times \lambda_5, \quad i = 1,2,3,\ldots,N, \quad j = 1,2,3,\ldots,D \tag{21}$$

Where $L_j$ and $U_j$ represent the lower and upper bound of $j_{th}$ variables, respectively. Besides, *D* shows the number of decision variables and $\lambda_5$ is a random number between 0 and 1. Finally, Equation (22) is used to model the screening process of GNDA, and Algorithm (2) represents the pseudo-code of GNDA.

$$X_i^{t+1} = \begin{cases} V_i^t, & \text{if } f(V_i^t) < f(X_i^t) \\ X_i^t, & \text{Otherwise} \end{cases} \tag{22}$$

**Algorithm 2.** Pseudo-code of GNDA

1.     Initialize the population by Equation (21)
2.       **While** (t < maximum number of iterations)
3.         Calculate the fitness of each population
4.         **for (i = 1: N)**
5.           Select a random number $\alpha$ between the range of [0, 1]
6.           **if ($\alpha$ > 0.5)** → Local Exploitation
7.             Select $X_{Best}^{t}$ and calculate **M** by Equation (17)
8.             Calculate **μ, δ**, and **η** by Equations (14) to (16), respectively
9.             Update the solutions by Equations (13) and (22)
10.             **else** → Global Exploration
11.             Update the solutions by Equations (18), (19), (20), and (22)
12.           **end if**
13.         **end for**
30.         t = t + 1
31.       **end while**
32.       **return $X_{Best}$**

### *3.4. Opposition-Based Learning Mechanism*

Opposition-Based Learning (OBL) is a straightforward search strategy, introduced by Tizhoos in 2005 [67]. It has been employed in MOAs to enhance the performance of original algorithms when searching for the optimal solution. Morevoer, it is a highly effective technique to avoid the local minima dilemma and accelerate the convergence rate. According to the OBL strategy, the fitness value of each solution and its corresponding opposite solution is calculated, then the best one is selected for the next phase. The OBL concept can be mathematically formulated as follow:

First, consider R as a real solution between the range of [$L_b$, $U_b$], where $L_b$ and $U_b$ are the lower and upper bounds of the search space. After that, the opposite solution of R is obtained using Equation (23).

$$R_{OBL} = L_b + U_b - R \tag{23}$$

Where $R_{OBL}$ represent the oppositional solution. For the N dimensional search spaces, the previous equation can be extended to:

$$R_{OBLi} = L_{bi} + U_{bi} - R_i, \quad i = 1,2,3,\ldots,N \tag{24}$$

Where $R_{OBLi} = [R_{OBL1} + R_{OBL2} + R_{OBL3} + \ldots + R_{OBLN}]$ and $R_i = [R_1 + R_2 + R_3 + \ldots + R_N]$ refer to the real solution and the oppositional solution in a N dimensional search space. Lastly, the algorithm updates the current optimal solution $R_i$ with the oppositional solution $R_{OBLi}$ if $f(R_{OBL}) > f(R)$.

## 4. Proposed Clustering Approach

In this section, a novel hybrid method based on Chimp Optimization Algorithm (ChOA) and Generalized Normal Distribution Algorithm (GNDA) with an Opposition-Based Learning (OBL) search strategy, entitled ChOAGNDA, is suggested for solving data clustering problems. The proposed approach is categorized into two subsections. Section 4.1 outlines the two proposed modifications into the ChOA algorithm. Following that, the improved version of ChOA is combined with GNDA and OBL techniques to form a new hybrid algorithm. Details of the proposed approach has been presented in Section 4.2.

### *4.1. ChOA Modifications*

The proposed modifications into the ChOA algorithm are divided to three parts. The first part discusses two versions of the ChOA clustering technique, while the second part describes the chaotic strategy in ChOA. The final section goes into the details of data clustering based on the ChOA algorithm.

### *4.1.1. Versioning Scheme*

As already mentioned, various types of chimps have different behaviors in their local and global searches. Therefore, each independent group employs its own strategy to update f. There are a variety of continuous functions that can be used to update f. The only characteristic of these functions is that the value of f must be declined after each iteration. Among all the available continuous functions, some of the most suitable ones are selected to propose two versions of clustering algorithms entitled ChOA(I) and ChOA(II). Table (2) demonstrates the dynamic coefficients of the f vector, where t and T are the number of current iteration and the maximum number of iterations, respectively. Further, the mathematical models of the proposed dynamic coefficients are illustrated in Figure (3).

**Table 2.** The dynamic coefficient of f vector.

| Groups | ChOA(I) | ChOA(II) |
| --- | --- | --- |
| Group 1 | $1.95 - 2\frac{t^{1/4}}{T^{1/3}}$ | $2.5 - 2\frac{\log(t)}{\log(T)}$ |
| Group 2 | $1.95 - 2\frac{t^{1/3}}{T^{1/4}}$ | $-2.2\frac{t^3}{T^3} + 2.5$ |
| Group 3 | $\left(-3\frac{t^{1/3}}{T^{1/3}}\right) + 1.5$ | $2.2 + 2\exp[-(4^{t}/_{T})^2]$ |
| Group 4 | $\left(-2\frac{t^3}{T^3}\right) + 1.5$ | $2.5 + 2(^{t}/_{T})^2 - 2(2^{t}/_{T})$ |

There are some potential advantages behind the dynamic coefficients, which help to enhance the performance of the proposed clustering algorithms. These points can be summarized as follows:

- Independent groups of chimps include Exponential and Logarithmic functions (non-linear functions) which can efficiently handle the complex clustering tasks.
- Each group of chimps has different abilities to discover the search space for finding the best cluster centers.
- Dynamic coefficient helps the clustering algorithm to achieve an appropriate balance between the local and global search.

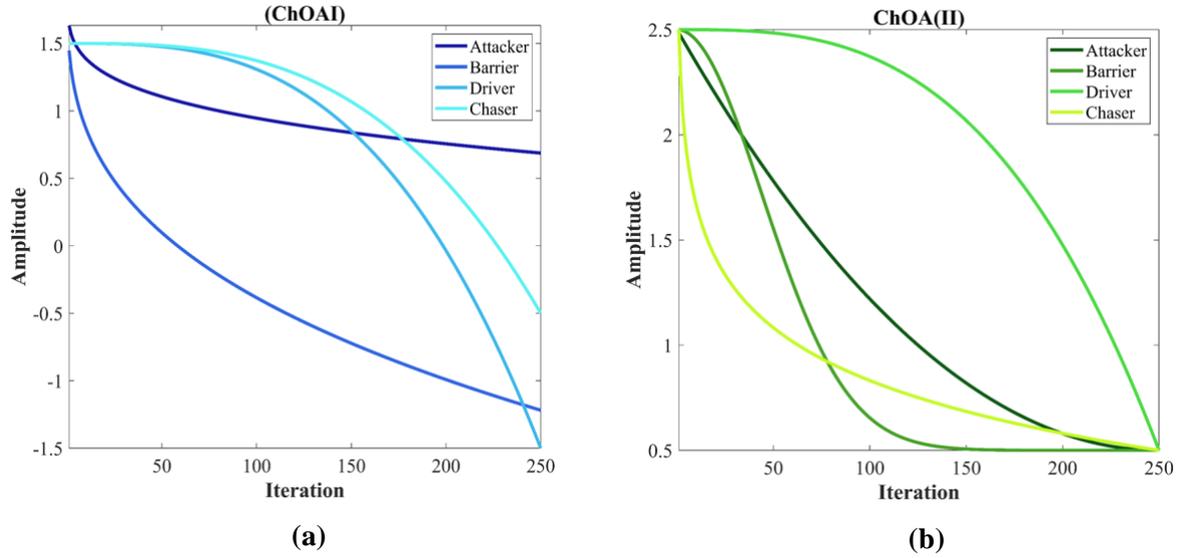

**Figure 3.** Mathematical models of the proposed dynamic coefficients for (a) ChOA(I) and (b) ChOA(II).

### 4.1.2. Chaotic Scheme

The chaotic value (M) is another significant parameter of ChOA that can be used to increase the performance of the proposed clustering algorithm. It should be noted that the remarkable improvement in the proposed algorithm has not been made only by grouping chimps into independent groups but also by using the new chaotic map in the final stage. In this paper, seven well-known chaotic maps [68] are employed to simulate the random behavior of the chimps at the last stage of the hunting process. The initial value is 0.7 for all the chaotic maps. It should be noted that the initial value of the chaotic maps can be any number in the interval of [0, 1]. Table (3) represents the details of the selected chaotic maps.

Generally, the following points can be considered to understand how chaotic maps affect the performance of ChOA clustering algorithm:

- Chaotic maps could promote the exploration and exploitation phase.
- Chaotic maps could be beneficial in solving the slow convergence rate issue.
- Chaotic maps help ChOA to escape from a local optima problem.
- Chaotic maps are effective in finding the best possible cluster center.
- Chaotic maps could be effective in achieving the minimum SICD result.
- Chaotic maps could decrease the error rate value.
- Chaotic maps do not need any additional computational cost for the proposed algorithm.

**Table 3.** Details of chaotic maps.

| Name | Chaotic map | Range |
|---|---|---|
| Circle | $X_{i+1} = \mod\left(X_i + b - \left(\frac{a}{2\pi}\right)\sin(2\pi X_k), 1\right)$, $a = 0.5$ and $b = 0.2$ | (0, 1) |
| Gauss/mouse | $X_{i+1} = \begin{cases} 1 & X_i = 0 \\ \frac{1}{\mod(X_i, 1)} & \text{otherwise} \end{cases}$ | (0, 1) |
| Logistic | $X_{i+1} = aX_i(1 - X_i)$, $a = 4$ | (0, 1) |
| Sine | $X_{i+1} = \frac{a}{4}\sin(\pi X_i)$, $a = 4$ | (0, 1) |
| Singer | $X_{i+1} = \mu(7.86X_i - 23.31X_i^2 + 28.75X_i^3 - 13.302875X_i^4)$, $\mu = 1.07$ | (0, 1) |
| Tent | $X_{i+1} = \begin{cases} \frac{X_i}{0.7} & X_i < 0.7 \\ \frac{10}{3}(1 - X_i) & X_i > 0.7 \end{cases}$ | (0, 1) |
| Piecewise | $X_{i+1} = \begin{cases} \frac{X_i}{P} & 0 \leq X_i < P \\ \frac{X_i - P}{0.5 - P} & P \leq X_i < 0.5 \\ \frac{1 - P - X_i}{0.5 - P} & 0.5 \leq X_i < 1 - P \\ \frac{1 - X_i}{P} & 1 - P \leq X_i < 1 \end{cases}$  $P = 0.4$ | (0, 1) |

In brief, the unconditional (chaotic) behavior in the final stage helps ChOA to improve the performance of the proposed clustering algorithm. Chaotic maps have very unique shapes, where they provide a large and extremely variable amplitude in the early stages, while their amplitude and variableness decline dramatically in the following stages. These special shapes of chaotic maps make chimps behave both very broadly in the early stages and narrowly in the final stages. In general, chaotic maps provide a smooth transition between global and local search capability. These maps help the proposed algorithm to escape from a local optimum problem due to the stochastic movement of chimps in the final stages. Accordingly, chimps are more likely to discover promising areas of search space and exploit the best possible cluster center. These are the primary reasons for the superior performance of ChOA in solving data clustering problems.

### *4.1.3. Description of ChOA Clustering Algorithm*

The main aim of the ChOA algorithm is to solve data clustering problems while mitigating the aforementioned drawbacks of the other approaches. in brief, the task of the proposed algorithm is to make an efficient partitioning of N data objects into K prespecified number of clusters. Data clustering based on SIAs consists of two major parts: defining an objective function and encoding candidate solutions (individuals). The objective function of the proposed clustering algorithm is fully discussed in the previous section. On the other hand, transforming the ChOA into a clustering-based algorithm requires the encoding of candidate solutions. For this purpose, Chimps are assigned to represent the clustering solutions, where each solution refers to the sets of cluster centroids. Attacker chimps are the best set of cluster centers that has the highest objective value, while driver chimps are the worst set of cluster centers with the lowest objective value. Lastly, the algorithm's objective is to improve the performance of the attacker chimps during each iteration. Figure (4) demonstrates a candidate solution with K clusters and d dimensions. Finally, Algorithm (3) describes the key steps of ChOA clustering algorithm in detail, and Figure (5) illustrates the flow chart of the proposed method.

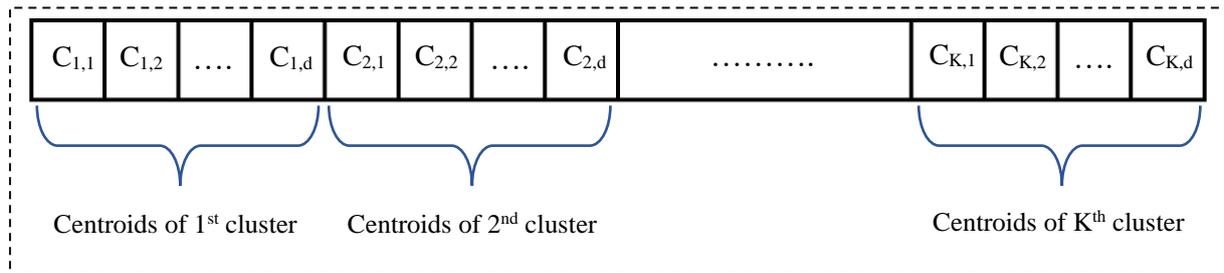

**Figure 4.** Encoding of a K×d candidate solution.

In brief, the ChOA clustering algorithm starts by creating a random population of chimps. Afterward, chimps are randomly categorized into four independent groups: attacker, barrier, chaser, and driver. Each independent group employs its own strategy to update its f coefficients. Then, the position and fitness value of each chimp (candidate solution) are calculated for the next stage. During the iteration process, each group of chimps estimates the prey's location and then updates their positions according to the potential location of prey. In the evaluation procedure, the previous best fitness value generated by ChOA (best solution) is compared to the new population. The chimp with the highest fitness value is taken to be considered as the best solution (cluster center). Similarly, the coefficient parameters (f, c, and M) are updated before being used to calculate a and d. The inequality $|a|<1$ obliges the chimps to move toward the prey, while inequality $|a|>1$ obliges the chimps to move backward the prey. Iteratively, the positions and fitness values of the new solutions are updated until the termination condition is reached. Eventually, the ChOA clustering algorithm selects $X_{Attacker}$ as the best set of cluster centers and assigns each data object to the nearest cluster center.

*4.2. Proposed ChOAGNDA Algorithm*

In this section, we described the key characteristics of the proposed hybrid approach, called ChOAGNDA, which is a selective opposition algorithm based on ChOA and GNDA algorithms. Although the proposed ChOA clustering algorithm is capable of dealing with various types of clustering problems, it still suffers from the following deficiencies: (i) slow convergence rate; (ii) getting trapped in local optima; (iii) failing to retain an appropriate balance between exploration and exploitation phases; and (iv) not always finding the best solution. To this aim, the best-obtained version of ChOA, entitled ChOA(II), is combined with GNDA algorithm and OBL technique to develop an efficient clustering approach. Figure (6) shows the overall framework of the proposed ChOAGNDA algorithm.

**Algorithm 3.** The procedure of ChOA clustering algorithm

1. Define objective function:
2. - $F_{obj}$: Sum of the Intra-Cluster Distances (SICD)
3. Set the initial parameters:
4. - **K**: The number of clusters
5. - $Iter_{max}$: Maximum number of iterations
6. - **P**: The size of population
7. - **I**: Input dataset
8. Initialize algorithm's parameters: **f**, **a**, **c**, and **M**
9. Initialize the position of chimps
10. Divide chimps randomly into different groups
11. Calculate the fitness of each chimp: $F_{Attacker}$, $F_{Chaser}$, $F_{Barrier}$, and $F_{Driver}$
12. Calculate the position of each chimp: $X_{Attacker}$, $X_{Chaser}$, $X_{Barrier}$, and $X_{Driver}$
13. Arrange the chimp's positions according to their fitness value:
14. - $X_{Attacker}$: The best set of cluster centers
15. - $X_{Chaser}$: The second-best set of cluster centers
16. - $X_{Barrier}$: The third-best set of cluster centers
17. - $X_{Driver}$: The fourth-best set of cluster centers
18. **While** (Iter < $Iter_{max}$)
19. Update population by new chimp:
20. Calculate the fitness value of updated chimp ($F_{new}$)
21. Update **Attackers**, **Barrier**, **Chaser**, and **Driver**:
22. - Update **Attacker**: If $F_{new} < F_{Attacker}$
23. - Update **Chaser**: If $F_{new} > F_{Attacker}$ && $F_{new} < F_{Chaser}$
24. - Update **Barrier**: If $F_{new} > F_{Attacker}$ && $F_{new} > F_{Chaser}$ && $F_{new} < F_{Barrier}$
25. - Update **Driver**: If $F_{new} > F_{Attacker}$ && $F_{new} > F_{Chaser}$ && $F_{new} > F_{Barrier}$ && $F_{new} > F_{Driver}$
26.     **for** each group of chimps
27.       Use its group strategy to update **f**, **c**, and **M**
28.       Use **f**, **c**, and M to calculate **a** and **d**
29.     **end for**
30.      **for** each search chimp
31.        **if** (|a| < 1)
32.        Calculate the position of updated chimps by Equation (4)
33.        **else if** (|a| > 1)
34.        Select a random search agent
35.        **end if**
36.      **end for**
37.     Update **f**, **a**, **c**, and **M**
38.     Update $X_{Attacker}$, $X_{Chaser}$, $X_{Barrier}$, and $X_{Driver}$
39.     Iter = Iter + 1
40.    **end while**
41.   **return** $X_{Attacker}$ as the best set of cluster centers
42. Set $X_{Attacker}$ as the best set of cluster centers
43. Assign each data object to the cluster that has the closest cluster center

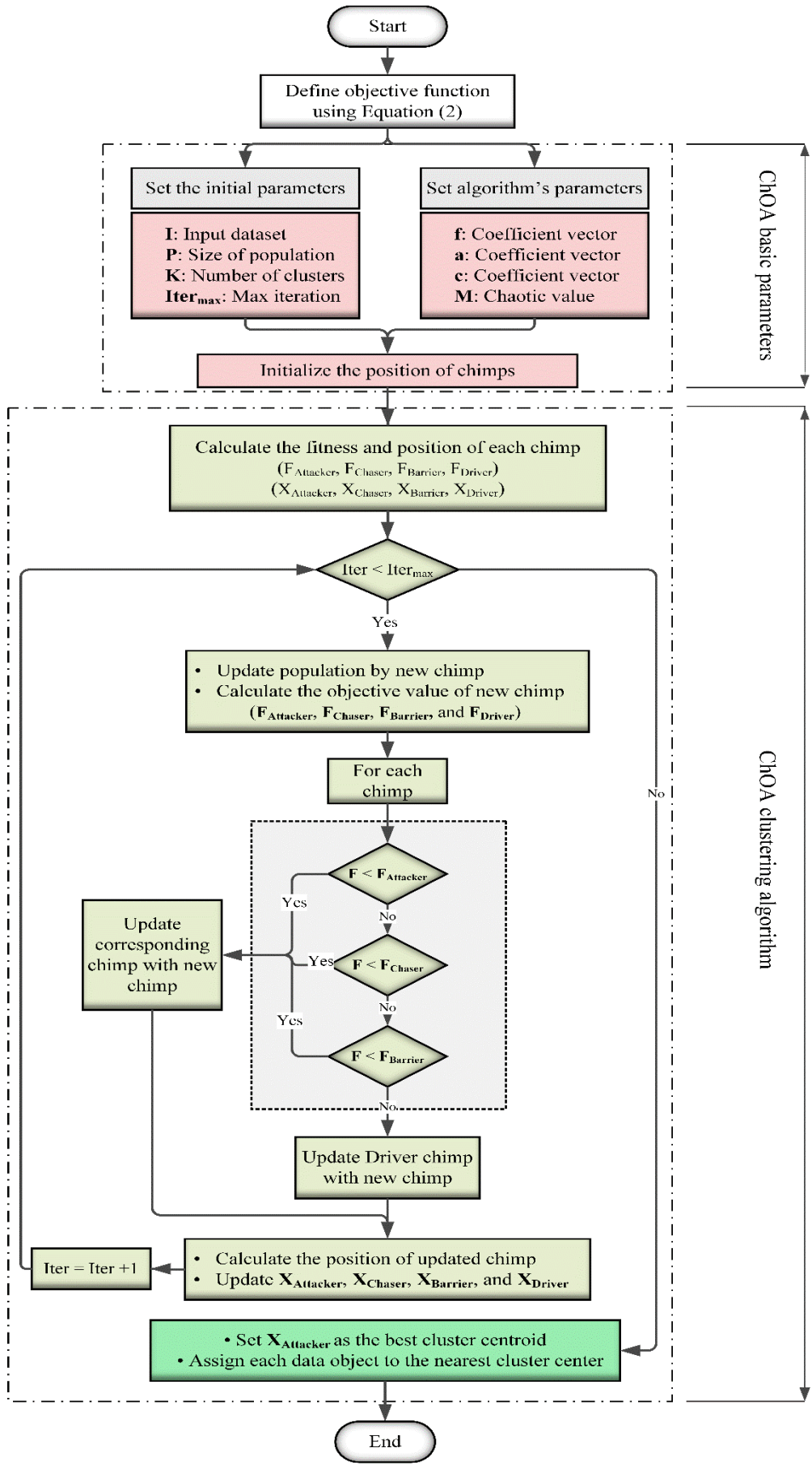

**Figure 5.** Flow chart of the proposed ChOA for data clustering.

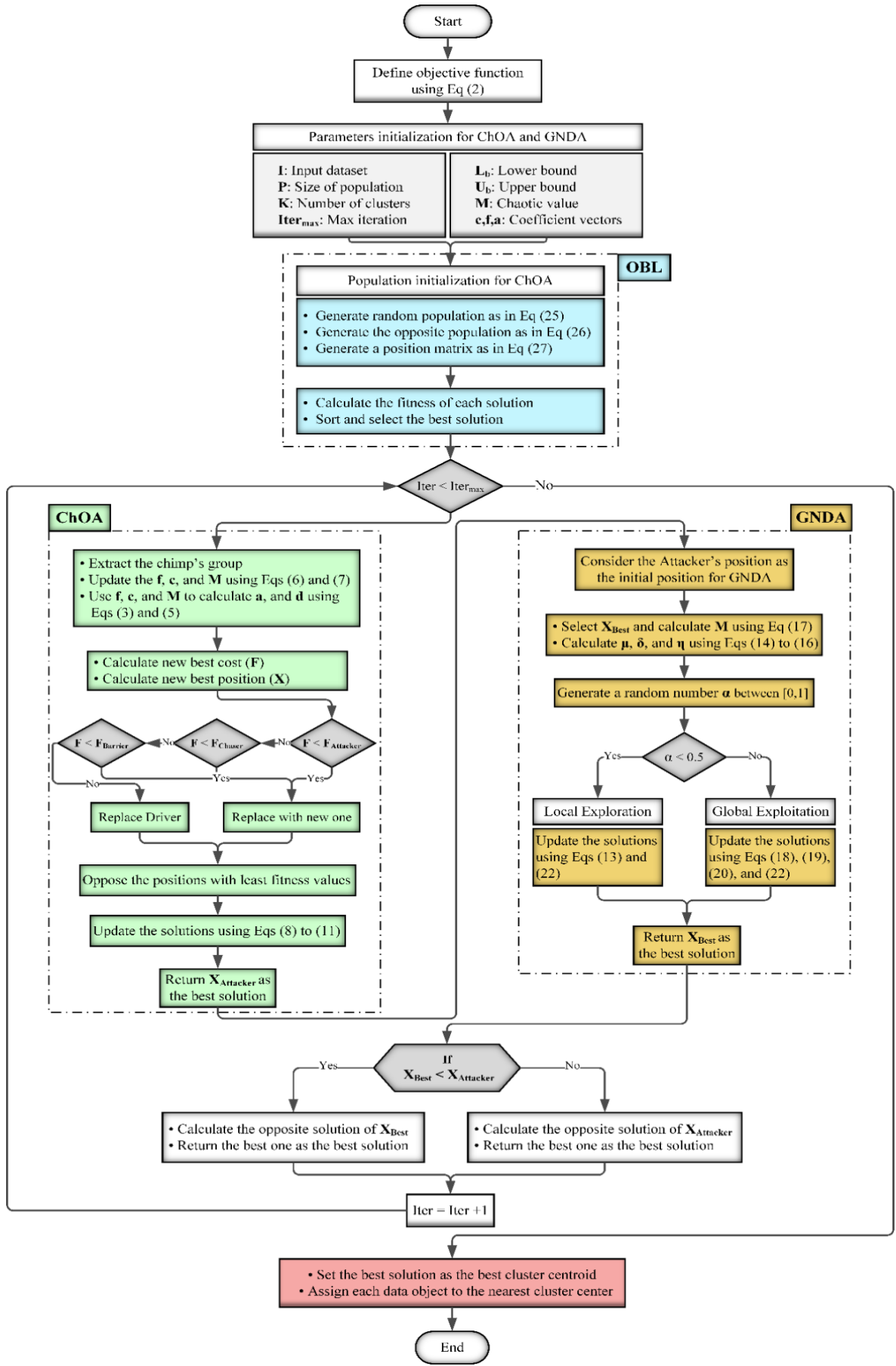

**Figure 6.** Flow chart of the proposed ChOAGNDA for data clustering.

Generally, the following steps are the main skeleton of the proposed ChOAGNDA algorithm for data clustering tasks. At first, the algorithm generates a random population as Equation (25), then calculates the opposite population as Equation (26), and finally sorts the population matrix as Equation (27).

$$\text{Population} = \begin{bmatrix} R_{1,1} & \cdots & R_{1,N} \\ \vdots & \ddots & \vdots \\ R_{\frac{P}{2},1} & \cdots & R_{\frac{P}{2},N} \end{bmatrix}_{\frac{P}{2} \times N} \tag{25}$$

$$\text{Opposite Population} = \begin{bmatrix} R_{OBL_{1,1}} & \cdots & R_{OBL_{1,N}} \\ \vdots & \ddots & \vdots \\ R_{OBL_{\frac{P}{2},1}} & \cdots & R_{OBL_{\frac{P}{2},N}} \end{bmatrix}_{\frac{P}{2} \times N} \tag{26}$$

$$\text{Position Matrix} = \begin{bmatrix} X_{1,1} & \cdots & X_{1,N} \\ \vdots & \ddots & \vdots \\ X_{P,1} & \cdots & X_{P,N} \end{bmatrix}_{P \times N} \tag{27}$$

Where P represents the number of population, N shows the dimension of each population, and X is the position of each population. Briefly, the ChOAGNDA clustering algorithm consists of five main phases. In the first phase, the proposed algorithm starts with a random population of chimps. To this aim, not only the initial population of search agents is generated using Equation (25) but also the opposite population is produced using Equation (26) according to the OBL technique. Following that, the fitness of each population is calculated, and the population with the highest fitness value is selected for the next phase. In the second phase, the ChOA algorithm is employed to discover the search space for finding the best possible solution. Accordingly, it returns the best solution as $X_{Attacker}$ and replaces the solutions with the lowest fitness value with their corresponding opposite solutions. Next, the GNDA algorithm begins the third phase with the solution produced by ChOA during the previous phase. Then, it tries to optimize the search space and returns $X_{Best}$ as the best optimal solution. In the fourth phase, $X_{Attacker}$ and $X_{Best}$ are compared, and the best one with the lowest fitness value is chosen for the following process. After that, the algorithm calculates the opposite solution of $X_{Attacker}$ or $X_{Best}$ to ensure that the optimal solution is achieved. Iteratively, the positions and fitness values of the new solutions are updated until the termination condition is reached. Lastly, the ChOAGNDA selects the best solution as the best set of cluster centers and assigns each object to the nearest cluster center.

## 5. Experimental Analysis and Discussion

This section investigates the performance of the proposed approaches in solving data clustering problems by applying them to the eight standard benchmark datasets. The results are compared against ten well-known meta-heuristic algorithms and five recently proposed clustering techniques. To this end, the entire algorithms are implemented using MATLAB 2018b, and all experiments are carried out with an Intel Core i5 processor, 2.4 GHz CPU, and 6 GB of RAM. The following subsections elaborate a full explanation of parameter tuning, dataset description, evaluation criteria, experimental results, and statistical analysis.

## 5.1. Parameters Tuning

Optimization algorithms are notably sensitive to the value of their control parameters. Generally, these parameters have a significant effect on the algorithm's ability to find optimal solutions. For this reason, tuning the parameters of metaheuristic algorithms is a crucial task that should be performed appropriately. In this paper, the proposed methods are compared with several state-of-art algorithms including GA [23], PSO [27], MVO [60], GWO [61], ABC [69], ACO [70], WOA [71], and GNDA [66]. Table (4) represents the details of the optimal parameters for the mentioned algorithms.

**Table 4.** Parameters setting of the proposed approach and other existing algorithms.

| Algorithm | Abbreviation | Parameter | Value |
|---|---|---|---|
| ChOA | $r_1, r_2$ | Random numbers | [0, 1] |
| | M | Chaotic value | [0, 1] |
| | f | Coefficient vector | Table (2) |
| | P | No. of chimps | 60 |
| GA | Pc | Crossover probability rate | 0.8 |
| | Pm | Mutation probability rate | 0.3 |
| | Mu | Mutation rate | 0.02 |
| | nPop | Population size | 60 |
| PSO | $r_1, r_2$ | Random number | [0, 1] |
| | $C_1$ | Cognitive constant | 2 |
| | $C_2$ | Social constant | 2 |
| | W | Local constant | 0.7 |
| | P | No. of particles | 60 |
| ABC | M | Modification rate | 0.4 |
| | a | Acceleration coefficient | 1 |
| | $\alpha$ | Step size | 0.0001 |
| | P | No. of bees | 60 |
| ACO | $\alpha$ | Pheromone weight | 1 |
| | $\beta$ | Heuristic Weight | 1 |
| | Rho | Evaporation rate | 0.05 |
| GNDA | A, B, $\lambda_1, \lambda_2$ | Random numbers | [0, 1] |
| | $\beta$ | Adjust parameter | 5 |
| | P | Population size | 60 |
| WOA | A | Coefficient vector | [-2, 2] |
| | C | Coefficient vector | [0, 2] |
| | P | No. of whales | 50 |
| GWO | A | Coefficient vector | [-2, 2] |
| | C | Coefficient vector | [0, 2] |
| | P | No. of wolves | 60 |
| MVO | $r_1, r_2, r_3$ | Random numbers | [0, 2] |
| | $W_{min}$ | Minimum wormhole | 0.2 |
| | $W_{max}$ | Maximum wormhole | 1 |
| | P | No. of search agents | 60 |

These experimental setups have been chosen according to the recommendations of the respective authors from previous studies. Additionally, each algorithm contains some common parameters such as population size, the number of iterations, and the number of independent runs. We set the same value for all the common parameters to ensure fair comparisons with other algorithms. This study considers the maximum number of 900 iterations for the UCI datasets and the maximum number of 200 iterations for the shape datasets. It should be noted that each experiment is executed 50 times independently to obtain more reliable and confident results.

*5.2. Dataset Description*

In this paper, ten benchmark datasets including five real-world datasets from the UCI data repository and three shape datasets from the previous literature [72–74], have been employed to assess the performance of the algorithms. The purpose of considering shape datasets is to evaluate the capability of the algorithms in solving complex shapes of clusters. The characteristics of these datasets are diverse in terms of the number of clusters (classes), features, and data samples. A full explanation about each UCI dataset is provided as follows:

- **Iris dataset:** this dataset, which was created by Fisher in 1936, is one of the most widely used datasets in data mining tasks such as classification, prediction, and clustering. Iris dataset includes four attributes: length of sepals, width of sepals, length of petals, and width of petals, as well as three classes namely Virginica, Versicolor, and Setosa, each with 50 samples.

- **Wine dataset:** this dataset illustrates the quality of three kinds of wines derived from various cultivars and grown in a particular region of Italy. The chemical examinations of the dataset are conducted by Forina et al. (1991). The Wine dataset contains 178 instances, 13 numerical attributes, and three classes, each with 59, 71, and 48 samples.

- **Cancer dataset:** Wisconsin Breast Cancer dataset was periodically collected by Wolberg from his clinical cases at the University of Wisconsin clinics in Madison, United States of America (USA). It represents whether the patients suffer from serious breast cancer or not. This dataset consists of 683 instances, nine attributes, and two classes: malignant and benign, with 239 and 444 samples, respectively.

- **Blood dataset:** this dataset was collected from 748 donors at the Blood Donor Center in Taiwan. It includes 748 data samples, where each sample is represented by four attributes entitled Frequency, Recency, Monetary, and Time. Moreover, the Blood dataset is a type of binary dataset in which the instances are divided into two classes: whether a donor donated blood in March 2007 or not.

- **CMC dataset:** Contraceptive Method Choice (CMC) dataset was extracted from the Indonesia Demographic and Health Survey. It contains information about the contraceptive method decisions of pregnant women. This dataset consists of 1473 data samples where each sample is expressed by nine attributes. CMC classes are divided into three classes: not use method, short-term method, long-term method with 629, 510, and 334 instances, respectively.

Table (5) describes general information about all the datasets discussed above, including the number of instances, features, and classes.

**Table 5.** Summary of five UCI datasets and three shape datasets.

| Category | Dataset | No. of classes | No. of features | No. of instances | Size of classes |
|---|---|---|---|---|---|
| UCI datasets | Iris | 3 | 4 | 150 | 50, 50, 50 |
| | Wine | 3 | 13 | 178 | 59, 71, 48 |
| | Cancer | 2 | 9 | 683 | 239, 444 |
| | Blood | 2 | 4 | 748 | 178, 570 |
| | CMC | 3 | 9 | 1473 | 629, 510, 334 |
| Shape datasets | Path-based | 3 | 2 | 300 | 110, 97, 93 |
| | Flame | 2 | 2 | 240 | 88, 152 |
| | Aggregation | 7 | 2 | 788 | 170, 34, 45, 102, 130, 273, 34 |

### 5.3. Evaluation Criteria

In this study, we evaluate the performance of clustering algorithms using the following metrics: Sum of Intra-Cluster Distances (SICD), convergence rate, and Error Rate (ER). The primary measurement is SICD, and it is calculated using Equation (2) as described in Section 3. Note that the minimum value of SICD represents a better quality of data clustering. The convergence comparison is the second measurement, where the clustering algorithms are compared in terms of their convergence speed to the optimal solution. The last measurement is ER which specifies the percentage of misclassified data objects, as formulated in Equation (28):

$$\text{ER} = \frac{\text{Number of misclassified data objects}}{\text{Total number of objects}} \times 100 \quad (28)$$

### 5.4. Results and Discussion

Experimental results are categorized into four major stages:

**Stage 1:** We investigate the impact of dynamic coefficients (f) and chaotic maps (M) on the effectiveness of the proposed clustering algorithm. The results related to ChOA(I) [33] and the proposed ChOA(II) are assessed based on the SICD results and convergence behaviors.

**Stage 2:** According to the previous section, the best-obtained version of ChOA is selected for the proposed hybrid algorithm and the resluts are compared against eight well-known clustering algorithms in terms of the SICD value, convergence rate, and ER value.

**Stage 3:** Then, a comparative analysis of the ChOAGNDA and other clustering algorithms from the past literature is performed according to their SICD and ER results.

**Stage 4:** Lastly, to emphasize the capability of the proposed algorithm, the ChOAGNDA along with eight existing clustering approaches are statistically analyzed through the obtained SICD and ER results.

The SICD and ER performances of all the previous comparisons are reported in terms of the following criteria: the best, the mean, the worst, and the Standard Deviation (STD) value of the solutions. Note that the results are obtained from 50 independent runs.

*5.4.1. Assessment of the influence of dynamic coefficients and chaotic maps*

In this section, we perform several experiments to determine the performance of two ChOA versions through their chaotic maps. To this end, the impact of dynamic coefficients (f) and chaotic maps (M) are examined on the efficiency of ChOA(I) and ChOA(II) in terms of their objective values (SICDs) and convergence rates. For all the following tables, the concepts of best, mean, worst, and STD are considered to show the ability of proposed algorithms in solving data clustering problems. Note that the bracketed numbers in the following tables show the ranks obtained by each chaotic map, which ranged from 1 (best result) to 7 (worst result). Table (6) illustrates the SICD results of ChOA(I) according to seven chaotic maps on eight datasets. It can be observed that the ChOA(I) through its particular chaotic map (Gauss), achieved the best SICD results in terms of the best, mean, worst, and STD in five out of eight datasets. Furthermore, to make a better comparison, we provide Figure (6) to analyze the average results of SICD obtained by ChOA(I). The lower average value of SICD indicates the better performance of the proposed algorithm for solving clustering problems. According to Figure (6), the ChOA(I) based on the Gauss map provided the best results for Iris, Cancer, Blood, CMC, and Aggregation datasets, whereas the Singer, Piecewise, and Circle maps obtained the best results just for one dataset including Wine, Path-Based, and Flame datasets, respectively. On the other hand, the same observation from Table (7) shows the superior results of ChOA(II) based on the Gauss map regarding the best, mean, worst, and STD in seven out of eight datasets. Figure (7) represents the average results of SICD criteria for ChOA(II). It is evident that the Gauss map has produced better results among all the proposed maps for all datasets except one dataset (namely: Iris). For the Iris dataset, the lowest SICD value is obtained by the Sine map.

By inspecting the numerical results in Tables (6) and (7), we can conclude that the ChOA(II) has attained the best SICD results in comparison with ChOA(I). This significant performance is due to two key reasons: (a) the mechanism of updating strategy and (b) the local search capability. To put it another way, independent groups of ChOA(II) are more likely than ChOA(I) to detect the optimal solution at the early iterations, which leads to better behavior in local search more than global search. In the case of the chaotic map, the superior results of ChOA(II) are achieved according to the Gauss map. The particular form of the Gauss map enables chimps to efficiently discover the search space in both the early and late phases of the hunting process. More precisely, the Gauss map has several advantages such as increasing the exploration performance in the initial stages, improving the exploitation capability in the final stages, and decreasing the possibility of trapping in local optima. Figure (8) demonstrates the convergence analysis of ChOA(I) and ChOA(II) based on their particular chaotic variant (namely: Gauss) on eight datasets. As we can observe from this graph, ChOA(II) provides a better convergence rate than ChOA(I) for all the datasets. It is worth mentioning that the ChOA(II) has not only started with the smallest objective value at the beginning steps of the optimization process but also converged towards the optimum solution faster than ChOA(I).

Eventually, it can be claimed that the ChOA(II) based on the Gauss map proves its significant performance in terms of various criteria. To this end, the combination of ChOA(II) and Gauss map has been chosen as the best-obtained model for the following experiments.

**Table 6.** SICD results of the ChOA(I) based on seven chaotic maps on eight datasets after 50 runs.

| Dataset | Measure | Circle | Gauss | Sine | Singer | Tent | Logistic | Piecewise |
|---|---|---|---|---|---|---|---|---|
| Iris | Best | 97.54851 | 96.85061 | 97.47174 | 97.09601 | 96.95510 | 97.50693 | 97.55666 |
| | Mean | 99.53009 | **97.83918** | 99.16945 | 98.36368 | 98.54192 | 99.07854 | 98.20342 |
| | Worst | 102.9539 | 99.07350 | 101.1746 | 99.52520 | 99.11267 | 99.78421 | 99.11149 |
| | STD | 2.396627 | 0.740934 | 1.513020 | 0.994551 | 0.889109 | 0.938750 | 0.815624 |
| | Rank | (7) | (1) | (6) | (3) | (4) | (5) | (2) |
| Wine | Best | 16421.780 | 16344.551 | 16390.535 | 16415.814 | 16389.271 | 16386.011 | 16426.570 |
| | Mean | 16687.438 | 16457.708 | 16582.723 | **16447.419** | 16470.143 | 16478.350 | 16746.898 |
| | Worst | 16968.950 | 16555.211 | 16645.152 | 16481.488 | 16622.753 | 16555.304 | 16999.464 |
| | STD | 245.22460 | 86.247403 | 128.36470 | 26.169251 | 111.38096 | 76.039951 | 293.92258 |
| | Rank | (6) | (2) | (5) | (1) | (3) | (4) | (7) |
| Cancer | Best | 2964.3887 | 2964.3881 | 2964.3890 | 2964.3888 | 2964.3890 | 2964.3893 | 2964.3894 |
| | Mean | 2964.3915 | **2964.3909** | 2964.3964 | 2964.3922 | 2964.3926 | 2964.3945 | 2964.3960 |
| | Worst | 2964.3945 | 2964.3953 | 2964.3944 | 2964.3988 | 2964.3971 | 2964.4058 | 2964.4048 |
| | STD | 0.0024953 | 0.0032766 | 0.0059039 | 0.0041324 | 0.0034895 | 0.0076441 | 0.0061371 |
| | Rank | (2) | (1) | (7) | (3) | (4) | (5) | (6) |
| Blood | Best | 407745.95 | 407731.83 | 407755.97 | 407745.72 | 407742.32 | 407741.05 | 407739.45 |
| | Mean | 407793.24 | **407748.72** | 407768.70 | 407758.77 | 407764.09 | 407770.09 | 407776.26 |
| | Worst | 407810.81 | 407779.58 | 407785.08 | 407909.61 | 407786.01 | 407801.53 | 407800.07 |
| | STD | 44.983510 | 12.638360 | 72.321510 | 16.382892 | 17.897610 | 29.126569 | 29.015712 |
| | Rank | (7) | (1) | (4) | (2) | (3) | (5) | (6) |
| CMC | Best | 5653.6428 | 5552.4410 | 5611.1899 | 5582.7723 | 5590.3029 | 5586.0711 | 5659.9392 |
| | Mean | 6143.1224 | **5638.5444** | 5746.4596 | 5933.8117 | 5710.7537 | 5677.9923 | 5951.3896 |
| | Worst | 7034.1194 | 5688.5781 | 6008.8822 | 70118.791 | 5904.9778 | 5798.1505 | 7025.8425 |
| | STD | 695.48406 | 49.062837 | 184.88288 | 610.24267 | 126.13409 | 110.95835 | 602.88203 |
| | Rank | (7) | (1) | (4) | (5) | (3) | (2) | (6) |
| Path-Based | Best | 1425.0005 | 1425.0601 | 1425.0836 | 1427.2101 | 1425.5378 | 1424.8671 | 1424.9418 |
| | Mean | 1426.6204 | 1426.3430 | 1426.5705 | 1428.4715 | 1426.4387 | 1427.6385 | **1425.9667** |
| | Worst | 1428.6208 | 1429.4342 | 1431.2255 | 1429.7494 | 1428.0156 | 1432.5778 | 1427.8864 |
| | STD | 1.3523846 | 1.7686005 | 2.6098421 | 1.1179014 | 0.9928647 | 3.0010539 | 1.1662687 |
| | Rank | (5) | (2) | (4) | (7) | (3) | (6) | (1) |
| Flame | Best | 769.97760 | 769.97580 | 769.96840 | 769.97481 | 769.97054 | 769.98391 | 769.97210 |
| | Mean | **769.98615** | 769.98872 | 769.98837 | 769.99145 | 770.01640 | 770.00780 | 769.99560 |
| | Worst | 770.00511 | 769.99781 | 770.01301 | 770.01881 | 770.10948 | 770.02351 | 770.01031 |
| | STD | 0.0129142 | 0.0096717 | 0.0190561 | 0.0205182 | 0.0630889 | 0.0184114 | 0.0170493 |
| | Rank | (1) | (3) | (2) | (4) | (7) | (6) | (5) |
| Aggregation | Best | 2934.2637 | 2816.6471 | 2934.6703 | 3009.1905 | 2854.2696 | 2958.3298 | 2996.2643 |
| | Mean | 3030.6645 | **2994.6958** | 3063.6589 | 3063.1798 | 3030.6644 | 2997.4797 | 3038.0770 |
| | Worst | 3137.3475 | 3040.1369 | 3139.0926 | 3120.5615 | 3114.9775 | 3113.9597 | 3090.5396 |
| | STD | 74.159450 | 34.146433 | 84.239170 | 47.461059 | 106.59080 | 72.777519 | 34.548081 |
| | Rank | (4) | (1) | (7) | (6) | (3) | (2) | (5) |
| **Average Ranking** | | **4.875** | **1.5** | **4.875** | **3.875** | **3.75** | **4.375** | **4.75** |

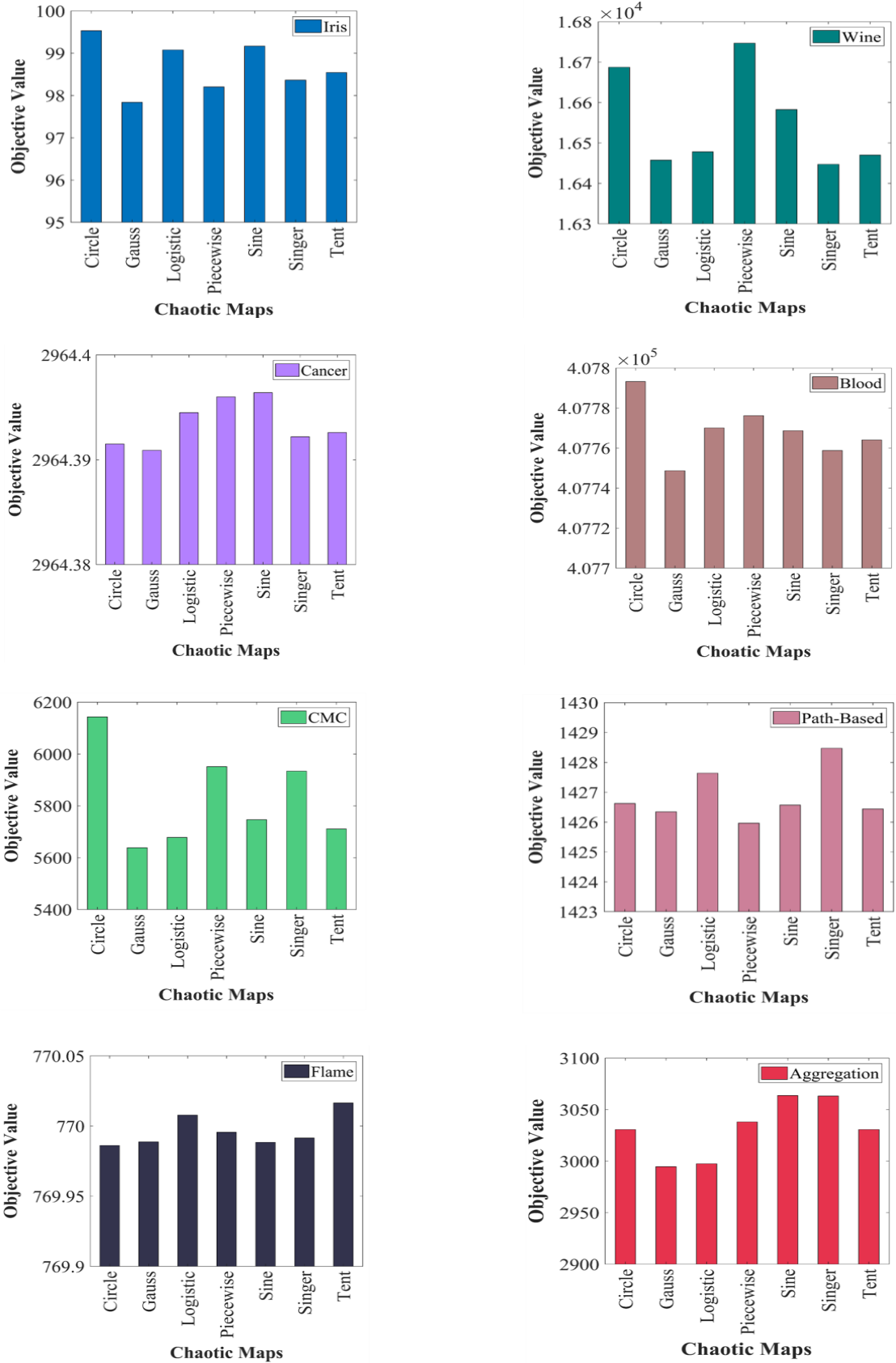

**Figure 6.** The average values of SICD for the ChOA(I) based on seven chaotic maps on eight datasets.

**Table 7.** SICD results of the ChOA(II) based on seven chaotic maps on eight datasets after 50 runs.

| Dataset | Measure | Circle | Gauss | Sine | Singer | Tent | Logistic | Piecewise |
|---|---|---|---|---|---|---|---|---|
| Iris | Best | 96.54991 | 96.54991 | 96.56020 | 96.56432 | 96.57611 | 96.58743 | 96.60101 |
| | Mean | 96.62142 | 96.58314 | **96.57033** | 96.57493 | 96.58900 | 96.60744 | 96.60843 |
| | Worst | 96.70700 | 96.60721 | 96.58000 | 96.59601 | 96.60658 | 96.62557 | 96.63377 |
| | STD | 0.059821 | 0.020715 | 0.009155 | 0.014363 | 0.011903 | 0.017246 | 0.014279 |
| | Rank | (7) | (3) | (1) | (2) | (4) | (5) | (6) |
| Wine | Best | 16308.784 | 16304.192 | 16307.092 | 16316.910 | 16307.684 | 16305.422 | 16307.510 |
| | Mean | 16310.383 | **16307.714** | 16313.324 | 16325.463 | 16311.013 | 16314.029 | 16313.913 |
| | Worst | 16338.189 | 16314.820 | 16318.662 | 16334.331 | 16320.509 | 16322.943 | 16331.174 |
| | STD | 12.822433 | 2.9196171 | 5.9706711 | 6.3406451 | 2.9996501 | 7.2810584 | 9.7545801 |
| | Rank | (2) | (1) | (4) | (7) | (3) | (6) | (5) |
| Cancer | Best | 2964.3888 | 2964.3861 | 2964.3886 | 2964.3885 | 2964.3885 | 2964.3888 | 2964.3883 |
| | Mean | 2964.3949 | **2964.3868** | 2964.3911 | 2964.3923 | 2964.3935 | 2964.3914 | 2964.3909 |
| | Worst | 2964.4001 | 2964.3902 | 2964.3936 | 2964.3968 | 2964.4035 | 2964.3945 | 2964.3971 |
| | STD | 0.0041823 | 0.0015770 | 0.0021377 | 0.0032692 | 0.0059823 | 0.0020440 | 0.0036251 |
| | Rank | (7) | (1) | (3) | (5) | (6) | (4) | (2) |
| Blood | Best | 407837.23 | 407714.23 | 407715.00 | 407715.14 | 407715.65 | 407715.38 | 407714.83 |
| | Mean | 407718.64 | **407714.23** | 407715.78 | 407715.60 | 407715.93 | 407716.52 | 407715.21 |
| | Worst | 408514.43 | 407714.23 | 407716.37 | 407715.86 | 407715.77 | 407717.50 | 407716.67 |
| | STD | 0.0005431 | 0.0000001 | 0.7002719 | 0.3309083 | 0.2584705 | 0.8914174 | 0.8373989 |
| | Rank | (7) | (1) | (4) | (3) | (5) | (6) | (2) |
| CMC | Best | 5537.1145 | 5534.0589 | 5534.9423 | 5535.8251 | 5537.2617 | 5538.0998 | 5535.2241 |
| | Mean | 5541.5880 | **5536.0522** | 5541.0747 | 5538.7952 | 5539.2564 | 5542.3312 | 5539.8356 |
| | Worst | 5545.8701 | 5539.8698 | 5550.3954 | 5544.5521 | 5542.1799 | 5544.8580 | 5544.4929 |
| | STD | 3.7953006 | 2.2840939 | 6.5772807 | 3.4396789 | 3.9414996 | 2.5477437 | 3.9379777 |
| | Rank | (6) | (1) | (5) | (2) | (3) | (7) | (4) |
| Path-Based | Best | 1424.9235 | 1424.8284 | 1425.3479 | 1425.1316 | 1425.2814 | 1424.9862 | 1425.1035 |
| | Mean | 1426.2359 | **1425.9166** | 1425.9238 | 1427.2298 | 1426.0102 | 1426.6745 | 1427.6521 |
| | Worst | 1427.1778 | 1426.7908 | 1426.8658 | 1431.0888 | 1426.8002 | 1429.9413 | 1429.8431 |
| | STD | 0.9629935 | 0.7976612 | 0.8043567 | 2.3106192 | 0.8026767 | 1.6964070 | 1.8613868 |
| | Rank | (4) | (1) | (2) | (6) | (3) | (5) | (7) |
| Flame | Best | 769.97171 | 769.96651 | 769.99181 | 769.98871 | 769.97755 | 769.96832 | 769.97644 |
| | Mean | 769.99194 | **769.96870** | 769.99349 | 769.99045 | 769.98175 | 769.98308 | 769.98478 |
| | Worst | 770.01930 | 769.96961 | 770.01834 | 770.01680 | 770.06581 | 770.00212 | 769.99933 |
| | STD | 0.0190969 | 0.0018207 | 0.0164117 | 0.0179179 | 0.0048283 | 0.0136659 | 0.0088264 |
| | Rank | (6) | (1) | (7) | (5) | (2) | (3) | (4) |
| Aggregation | Best | 2774.0879 | 2766.3764 | 2779.4178 | 2767.6853 | 2751.4661 | 2780.6119 | 2781.5202 |
| | Mean | 2784.9206 | **2778.0064** | 2856.1863 | 2898.3827 | 2791.7621 | 2919.3303 | 2945.1331 |
| | Worst | 2843.6293 | 2786.4754 | 3023.1693 | 3075.4964 | 2818.9691 | 3050.0264 | 3205.8344 |
| | STD | 40.644002 | 7.9005134 | 100.74101 | 145.35414 | 28.183541 | 88.913061 | 204.14626 |
| | Rank | (2) | (1) | (4) | (5) | (3) | (6) | (7) |
| **Average Ranking** | | **5.125** | **1.25** | **3.75** | **4.375** | **3.625** | **5.25** | **4.625** |

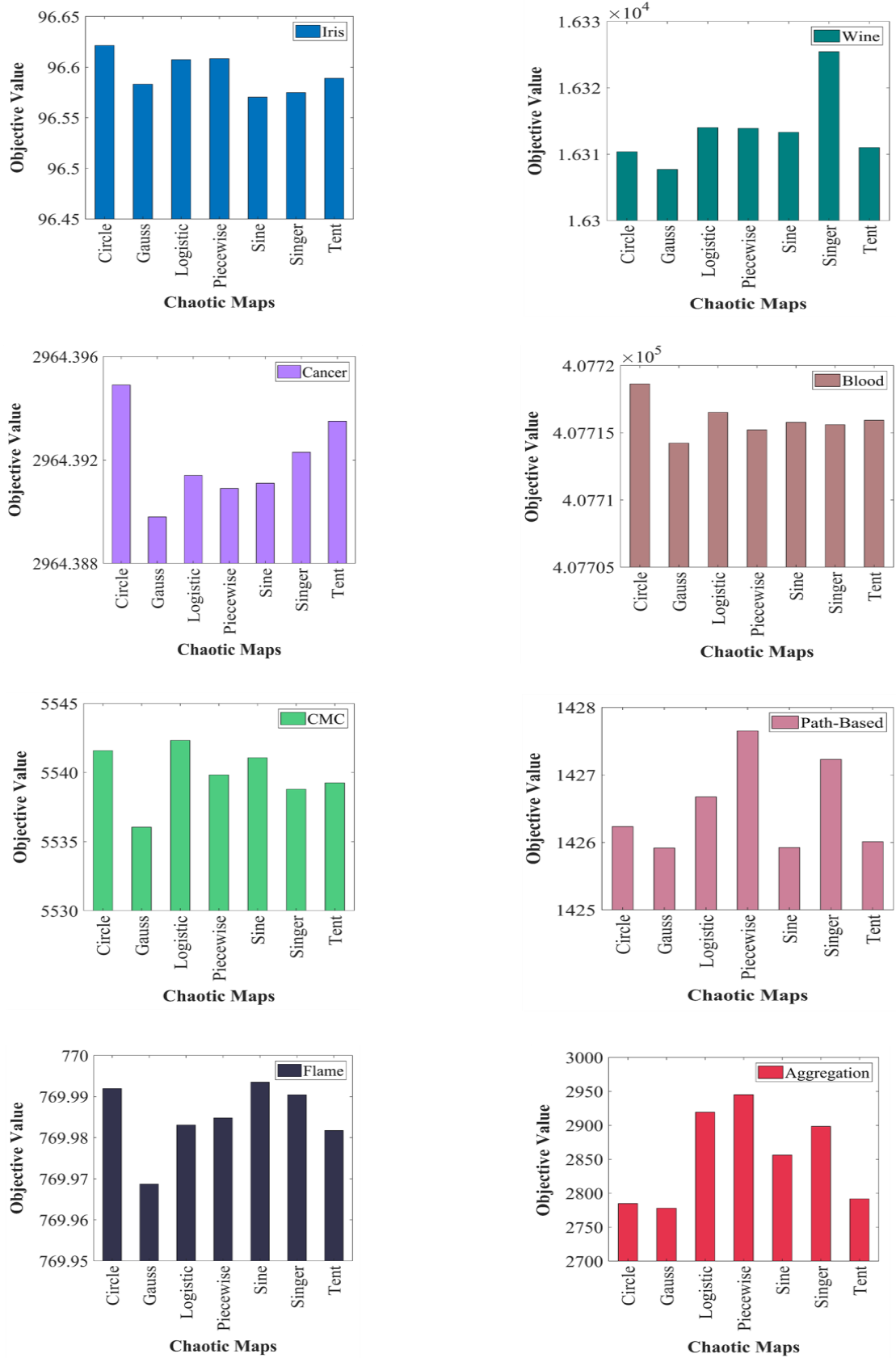

**Figure 7.** The average values of SICD for the ChOA(II) based on seven chaotic maps on eight datasets.

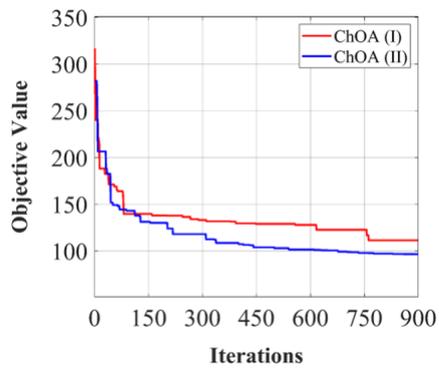
(a) Iris

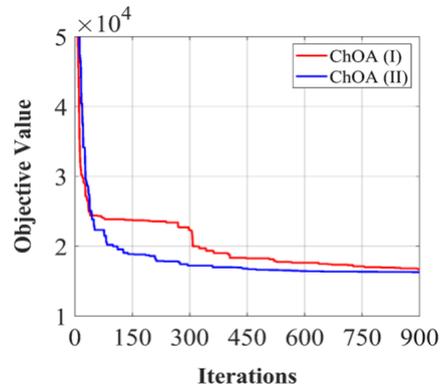
(b) Wine

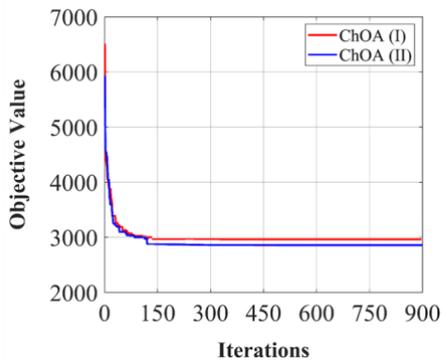
(c) Cancer

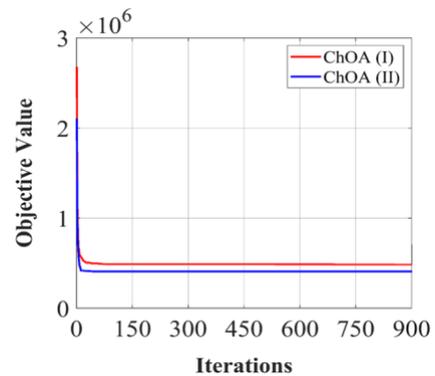
(d) Blood

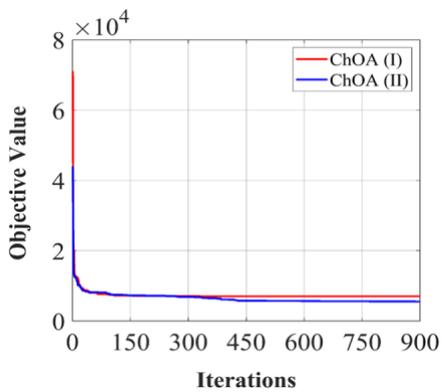
(e) CMC

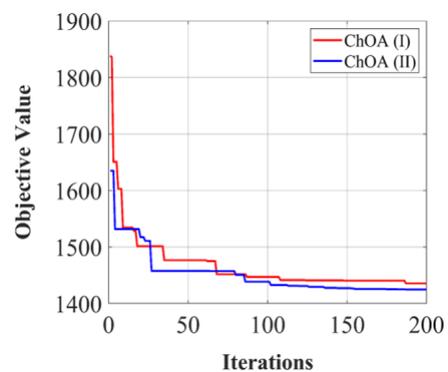
(f) Path-Based

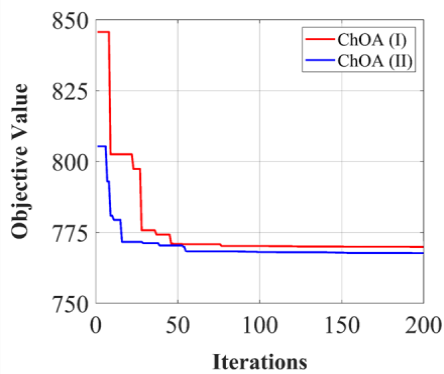
(g) Flame

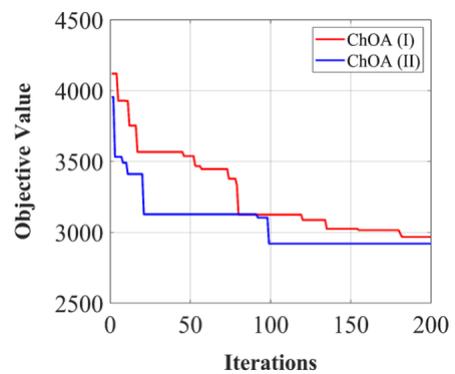
(h) Aggregation

**Figure 8.** The comparative convergence curves of ChOA(I) and ChOA(II) according to their best chaotic maps on (a) Iris, (b) Wine, (c) Cancer, (d) Blood, (e) CMC, (f) Path-Based, (g) Flame, and (h) Aggregation datasets.

*5.4.2. Comparison between ChOAGNDA and other meta-heuristic algorithms*

In this section, we verify the performance of the proposed approach (ChOAGNDA) on data clustering problems by comparing it against the most well-known meta-heuristic algorithms (K-Means, GA, PSO, MVO, GWO, ABC, ACO, GNDA, ChOA, and WOA). The algorithms are evaluated regarding SICD value, ER result, and convergence rate on eight clustering datasets. Note that the bracketed numbers in the following tables show the ranks obtained by each algorithm, which ranged from 1 (best result) to 11 (worst result).

- **SICD analysis:** Table (8) represents a comparison between ChOAGNDA and other clustering algorithms with respect to the SICD results on eight datasets. It is observed that ChOAGNDA significantly outperforms other algorithms in terms of the best, mean, worst, and STD in all datasets. Moreover, the minimum STD value of ChOAGNDA proves its satisfactory stability and robustness in achieving promising results.

- **Convergence analysis:** Figure (9) demonstrates the average convergence behaviors of the clustering algorithms regarding their objective values for eight datasets. It can be observed that the ChOAGNDA has a better convergence curve in seven out of eight datasets. In the case of the Iris dataset, ChOAGNDA, GA, MVO, and PSO algorithms begin the optimization with almost similar convergence trends, but ChOAGNDA beats all the other algorithms after a limited number of iterations. For the Wine dataset, WOA, ChOA, PSO, GWO, and MVO algorithms have the same convergence curve through the whole optimization process. On the other side, all of the examined algorithms perform a similar pattern on Cancer, Blood, and CMC datasets, where ChOAGNDA obtains the fastest trend and ranks first among all approaches. In the case of the Path-Based dataset, ChOAGNDA and GA start the optimization process far better than the other algorithms, and the ChOAGNDA eventually obtains the lowest objective value at the end of the process. For the Flame dataset, MVO generates a poor convergence rate compared to the other algorithms, while ChOAGNDA and PSO have the best convergence behavior and reach the optimal solution with the minimum number of iterations. For the Aggregation dataset, although PSO, MVO, GA, GWO, and WOA perform a better convergence trend at the initial steps of the optimization process, ChOAGNDA outperforms all existing approaches after passing half of the total iterations. Additionally, Figure (10) provides a more detailed comparison of the proposed algorithm the original algorithms used in our approach. It is obvious that the proposed ChOAGNDA not only converges to the optimal solution during the initial iterations but also achieves the minimum SICD value for all the tested datasets.

- **ER analysis:** Table (9) illustrates a comparison of ChOAGNDA with other meta-heuristic algorithms in terms of the ER results on eight datasets. According to this table, the minimum ER value is obtained by the proposed clustering approach as 9.10% in Iris, 26.51% in Wine, 3.53% in Cancer, 34.81% in Blood, 49.33% in CMC, 24.70% in Path, 15.21 in Flame, and 23.09% in Aggregation dataset. These statistical results prove the superiority of ChOAGNDA over the compared algorithms on all datasets.

**Table 8.** The comparative results of ChOAGNDA and other clustering algorithms regarding the SICD value on eight datasets after 50 runs.

| Algorithm | Measure | Iris | Wine | Cancer | Blood | CMC | Path | Flame | Aggregation |
|---|---|---|---|---|---|---|---|---|---|
| K-Means | Best | 97.3243 | 16,55.6723 | 2986.9521 | 408336.41 | 5691.6285 | 1527.4675 | 781.2574 | 3352.5869 |
| | Mean | 104.7182 | 16963.0492 | 3032.2478 | 410298.29 | 5703.5253 | 1696.7224 | 818.5320 | 3635.1877 |
| | Worst | 124.2111 | 23755.0410 | 5216.0894 | 415329.73 | 5706.3197 | 1821.5590 | 853.4439 | 3931.2492 |
| | STD | 12.3875 | 1180.6942 | 315.1456 | 2436.52 | 2.5238 | 77.2663 | 28.3541 | 143.4151 |
| | Rank | (11) | (11) | (6) | (10) | (8) | (11) | (11) | (11) |
| GA | Best | 98.3741 | 16350.8024 | 2981.1355 | 408174.38 | 5588.7356 | 1442.8257 | 770.9752 | 2792.1168 |
| | Mean | 99.4470 | 16520.1760 | 3048.7738 | 411779.55 | 5632.2107 | 1457.4935 | 777.8678 | 2822.7590 |
| | Worst | 103.7833 | 16530.5729 | 3138.3160 | 416711.92 | 5739.6956 | 1613.2532 | 812.5009 | 3239.7090 |
| | STD | 15.3671 | 58.2548 | 49.8137 | 2723.58 | 74.1708 | 17.9593 | 9.1533 | 44.2162 |
| | Rank | (10) | (7) | (11) | (11) | (6) | (7) | (6) | (4) |
| PSO | Best | 96.8753 | 16305.4821 | 2975.3709 | 407853.43 | 5539.2834 | 1425.6268 | 769.9773 | 2763.1175 |
| | Mean | 98.1534 | 16317.2787 | 2982.6757 | 407997.81 | 5547.8037 | 1443.4412 | 770.4814 | 2879.0641 |
| | Worst | 99.7706 | 16344.7893 | 3054.5025 | 408339.55 | 5560.6549 | 1588.2870 | 832.1860 | 3888.4430 |
| | STD | 0.8530 | 13.6027 | 10.4675 | 128.92 | 7.4673 | 14.8569 | 0.6341 | 52.5715 |
| | Rank | (7) | (5) | (5) | (9) | (5) | (4) | (3) | (6) |
| MVO | Best | 96.6831 | 16319.2761 | 2964.7511 | 407725.35 | 5754.1564 | 1458.3487 | 770.4246 | 3043.7696 |
| | Mean | 98.5849 | 16376.7196 | 2965.2314 | 407816.54 | 5824.1445 | 1546.3827 | 813.8370 | 3441.3369 |
| | Worst | 99.7281 | 16422.6621 | 2965.9140 | 407875.90 | 5940.6081 | 1673.8310 | 864.5372 | 3819.7929 |
| | STD | 0.91289 | 20.2831 | 0.3249 | 81.3497 | 49.2360 | 45.6529 | 24.1463 | 151.4479 |
| | Rank | (8) | (6) | (4) | (8) | (11) | (10) | (10) | (10) |
| GWO | Best | 96.6671 | 16307.4370 | 2964.3901 | 407723.25 | 5538.3369 | 1427.5647 | 769.9772 | 2829.2701 |
| | Mean | 99.2368 | 16316.7419 | 2964.3949 | 407781.39 | 5544.6011 | 1465.6109 | 775.3835 | 2865.3370 |
| | Worst | 104.5501 | 16386.3261 | 2964.3991 | 407812.40 | 5549.3044 | 1499.7663 | 789.3884 | 2897.7719 |
| | STD | 13.2762 | 10.4675 | 0.0024 | 31.4540 | 4.1229 | 21.3615 | 3.7673 | 17.5243 |
| | Rank | (9) | (4) | (3) | (7) | (4) | (9) | (5) | (5) |
| ABC | Best | 97.0521 | 16431.2519 | 2988.4301 | 407722.98 | 5644.3381 | 1432.0479 | 769.9330 | 2897.3401 |
| | Mean | 97.9335 | 16844.6520 | 3041.8441 | 407750.51 | 5692.6004 | 1461.6157 | 778.3412 | 3087.1705 |
| | Worst | 99.1932 | 17225.7713 | 3124.4719 | 407782.28 | 5757.8220 | 1499.6310 | 794.9007 | 3223.5089 |
| | STD | 0.5412 | 288.7715 | 49.5219 | 23.2710 | 31.1663 | 10.2712 | 13.6561 | 20.5471 |
| | Rank | (6) | (10) | (10) | (6) | (7) | (8) | (7) | (7) |
| ACO | Best | 97.1241 | 16523.5298 | 2973.4419 | 407724.41 | 5715.8133 | 1439.2568 | 769.9988 | 2899.2945 |
| | Mean | 97.1709 | 16636.2372 | 3035.2670 | 407740.11 | 5789.6272 | 1457.2687 | 779.3134 | 3092.1786 |
| | Worst | 97.6729 | 16705.6603 | 3128.2662 | 407778.34 | 5820.3348 | 1490.5423 | 799.2086 | 3273.3245 |
| | STD | 0.5173 | 8.59156 | 21.3186 | 17.1065 | 39.2976 | 19.3714 | 7.4512 | 20.2208 |
| | Rank | (5) | (9) | (8) | (5) | (10) | (6) | (9) | (9) |
| WOA | Best | 96.7208 | 16291.9083 | 2994.4401 | 407720.33 | 5537.7530 | 1424.8395 | 769.9670 | 2782.7009 |
| | Mean | 96.7931 | 16295.1044 | 3036.1273 | 407727.45 | 5539.6891 | 1438.6106 | 770.5922 | 2798.3320 |
| | Worst | 96.8303 | 16320.7573 | 3081.4401 | 407751.90 | 5546.0744 | 1487.2219 | 771.2871 | 2817.9106 |
| | STD | 0.1081 | 2.7155 | 12.2019 | 3.5636 | 3.7991 | 13.4105 | 0.5130 | 8.3309 |
| | Rank | (3) | (2) | (9) | (3) | (3) | (3) | (4) | (3) |
| GNDA | Best | 97.1011 | 16521.4197 | 2972.3379 | 407722.41 | 5713.7103 | 1438.2108 | 769.9916 | 2898.1422 |
| | Mean | 97.1512 | 16633.1366 | 3033.1640 | 407739.09 | 5787.5357 | 1456.2200 | 779.3110 | 3091.1266 |
| | Worst | 97.6169 | 16703.5501 | 3126.1772 | 407775.55 | 5819.2205 | 1489.7323 | 799.2001 | 3271.4073 |
| | STD | 0.4183 | 7.4805 | 20.3186 | 16.9033 | 37.1976 | 19.3714 | 6.5612 | 18.2208 |
| | Rank | (4) | (8) | (7) | (4) | (9) | (5) | (8) | (8) |
| ChOA | Best | 96.5499 | 16304.1921 | 2964.3861 | 407714.23 | 5534.0589 | 1424.8284 | 769.9665 | 2766.3764 |
| | Mean | 96.5831 | 16307.7144 | 2964.3868 | 407714.23 | 5536.0522 | 1425.9166 | 769.9687 | 2778.0064 |
| | Worst | 96.6072 | 16314.8209 | 2964.3902 | 407714.23 | 5539.8698 | 1426.7908 | 769.9696 | 2786.4754 |
| | STD | 0.0207 | 2.9196 | 0.0015 | 0.0001 | 2.2840 | 0.7976 | 0.0010 | 7.9005 |
| | Rank | (2) | (3) | (2) | (2) | (2) | (2) | (2) | (2) |
| **ChOAGNDA** | **Best** | **96.5400** | **16292.1842** | **2964.3860** | **407714.19** | **5532.0156** | **1424.7147** | **769.9662** | **2712.9089** |
| | **Mean** | **96.5403** | **16292.1846** | **2964.3862** | **407714.21** | **5532.1847** | **1424.7166** | **769.9663** | **2713.1577** |
| | **Worst** | **96.5404** | **16292.6127** | **2964.3870** | **407714.28** | **5532.1893** | **1424.7958** | **769.9664** | **2734.4129** |
| | **STD** | **0.0001** | **0.0008** | **0.0002** | **0.0001** | **0.0007** | **0.0030** | **0.0001** | **0.400** |
| | **Rank** | **(1)** | **(1)** | **(1)** | **(1)** | **(1)** | **(1)** | **(1)** | **(1)** |

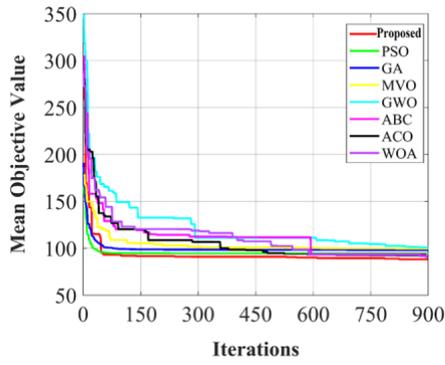
(a) Iris

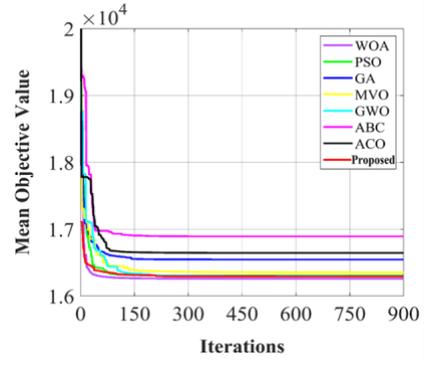
(b) Wine

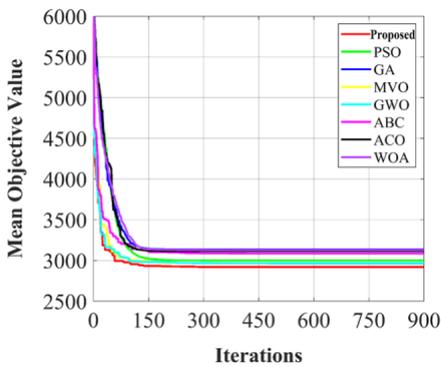
(c) Cancer

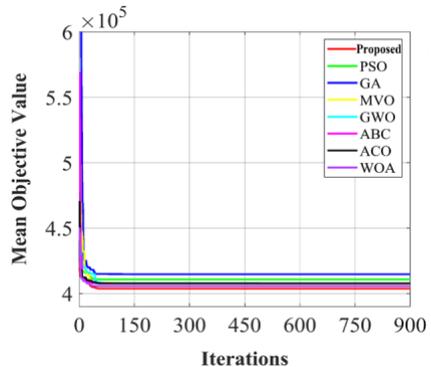
(d) Blood

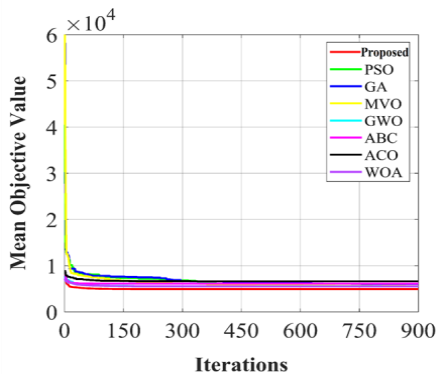
(e) CMC

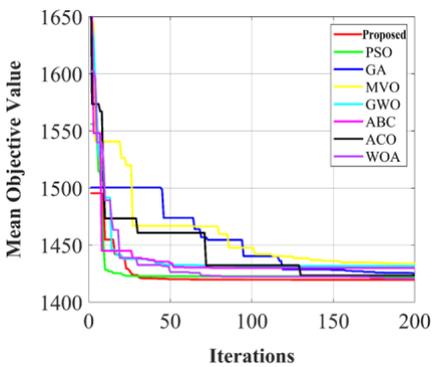
(f) Path-Based

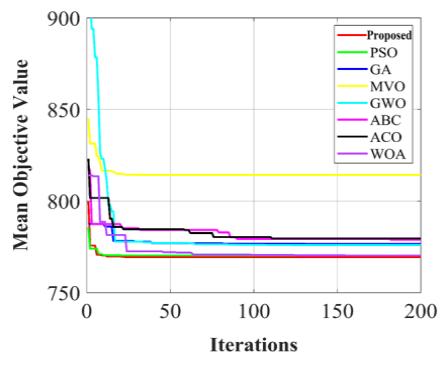
(g) Flame

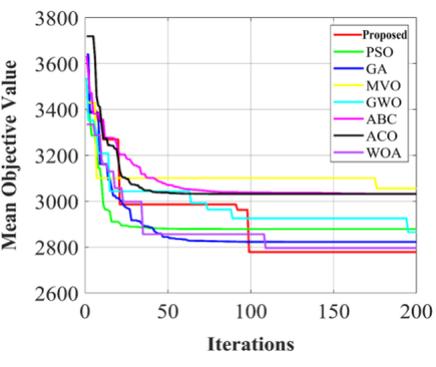
(h) Aggregation

**Figure 9.** The average convergence curves of ChOAGNDA and other state-of-art algorithms regarding the mean objective value on (a) Iris, (b) Wine, (c) Cancer, (d) Blood, (e) CMC, (f) Path-Based, (g) Flame, and, (h) Aggregation datasets.

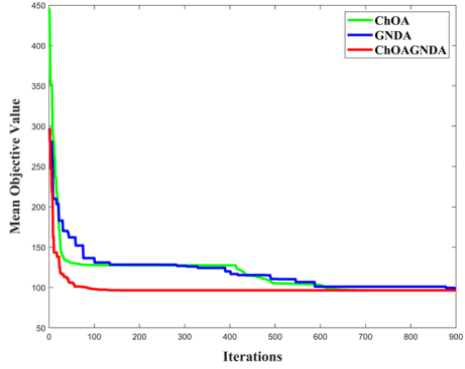
(a) Iris

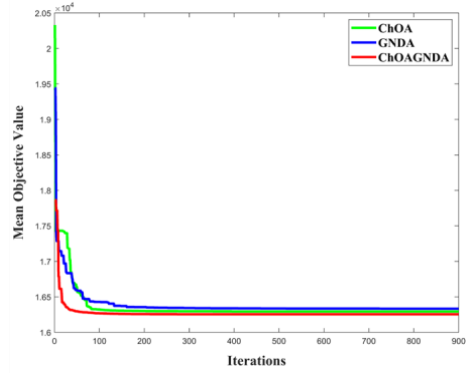
(b) Wine

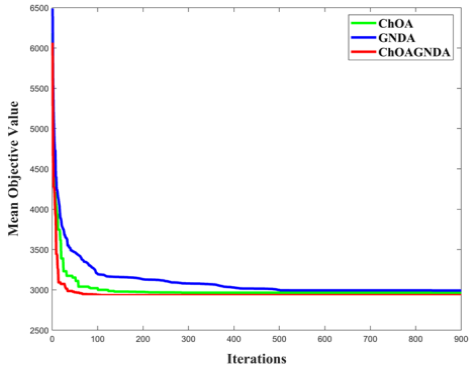
(c) Cancer

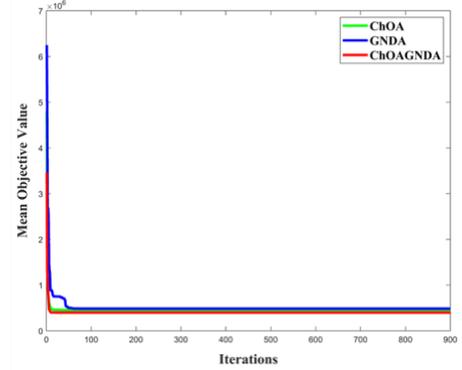
(d) Blood

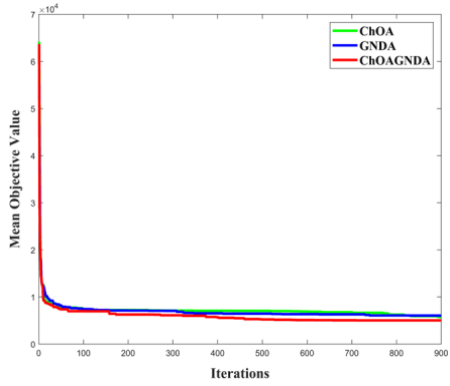
(e) CMC

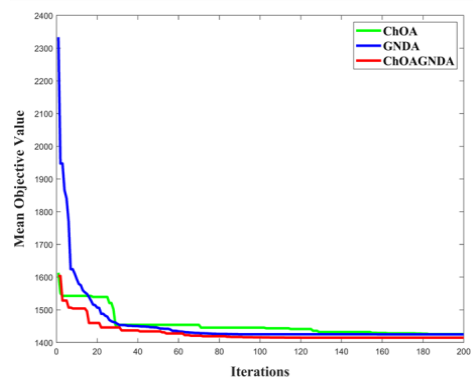
(f) Path-Based

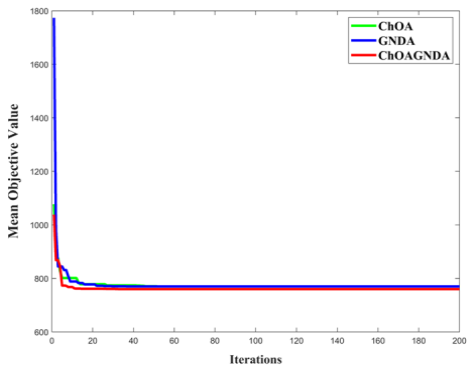
(g) Flame

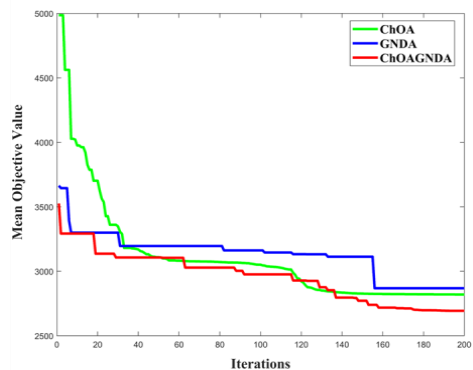
(h) Aggregation

**Figure 10.** The comparative convergence curves of ChOAGNDA, ChOA, and GNDA according to the mean objective value on (a) Iris, (b) Wine, (c) Cancer, (d) Blood, (e) CMC, (f) Path-Based, (g) Flame, and (h) Aggregation datasets.

**Table 9.** The comparative results of ChOAGNDA and other clustering algorithms regarding the average error rate (%) on eight datasets after 50 runs.

| Algorithm | Measure | Iris | Wine | Cancer | Blood | CMC | Path | Flame | Aggregation |
|---|---|---|---|---|---|---|---|---|---|
| K-Means | Mean (%) | 13.44 | 31.12 | 4.37 | 35.22 | 54.45 | 29.67 | 17.82 | 29.22 |
|  | Rank | (11) | (11) | (11) | (6) | (5) | (11) | (11) | (11) |
| GA | Mean (%) | 10.00 | 28.75 | 3.87 | 34.92 | 57.66 | 27.45 | 16.35 | 25.50 |
|  | Rank | (3) | (4) | (6) | (5) | (11) | (6) | (6) | (4) |
| PSO | Mean (%) | 10.06 | 28.79 | 3.79 | 34.90 | 54.48 | 26.57 | 15.92 | 25.88 |
|  | Rank | (6) | (7) | (4) | (3) | (7) | (4) | (3) | (6) |
| MVO | Mean (%) | 10.45 | 28.77 | 3.80 | 34.91 | 54.92 | 29.57 | 17.50 | 26.87 |
|  | Rank | (9) | (5) | (5) | (4) | (9) | (10) | (10) | (10) |
| GWO | Mean (%) | 10.73 | 29.48 | 3.65 | 34.90 | 55.23 | 29.34 | 16.03 | 25.73 |
|  | Rank | (10) | (10) | (3) | (3) | (10) | (8) | (5) | (5) |
| ABC | Mean (%) | 10.04 | 28.92 | 3.93 | 34.91 | 54.39 | 29.07 | 17.22 | 26.45 |
|  | Rank | (5) | (9) | (10) | (4) | (3) | (7) | (7) | (7) |
| ACO | Mean (%) | 10.02 | 28.78 | 3.88 | 34.90 | 54.47 | 26.62 | 17.38 | 26.80 |
|  | Rank | (4) | (6) | (8) | (3) | (6) | (5) | (8) | (8) |
| WOA | Mean (%) | 10.08 | 28.12 | 3.91 | 34.89 | 54.41 | 26.51 | 15.97 | 25.37 |
|  | Rank | (7) | (3) | (9) | (2) | (4) | (3) | (4) | (3) |
| GNDA | Mean (%) | 10.40 | 28.90 | 3.85 | 34.91 | 54.57 | 29.49 | 17.44 | 26.85 |
|  | Rank | (8) | (8) | (7) | (4) | (8) | (9) | (9) | (9) |
| ChOA | Mean (%) | 9.30 | 27.92 | 3.62 | 34.89 | 53.93 | 26.00 | 15.83 | 24.59 |
|  | Rank | (2) | (2) | (2) | (2) | (2) | (2) | (2) | (2) |
| **ChOAGNDA** | **Mean (%)** | **9.10** | **26.51** | **3.53** | **34.81** | **49.33** | **24.70** | **15.21** | **22.09** |
|  | **Rank** | **(1)** | **(1)** | **(1)** | **(1)** | **(1)** | **(1)** | **(1)** | **(1)** |

### 5.4.3. Comparison of the proposed method with past literature

In recent years, many studies have been focused on solving real-world optimization problems, particularly in the field of data clustering. To ensure the strength of our method, this section provides a comparison between the proposed algorithm (ChOAGNDA) and the most powerful metaheuristics-based clustering algorithms in previous literature. The results of the Gravitational Search Algorithm (GSA) [41], Gravitational Search Algorithm based on K-Means (GSA-KM) [75], Improved Krill Herd (IKH) [76], the combination of Improved Cuckoo optimization with Modified Particle Swarm Optimization and K-Harmonic Means (ICMPKHM) [77], Symbiotic Organism Search (SOS) [78], Quantum-inspired Ant Lion Optimized hybrid K-means (QALO-K) [79], and Hybrid Harris Hawks Optimization (H-HHO) (Abualigah et al., 2021) are directly taken from their references, while the results of the Cuckoo Search (CS), Genetic Quantum Cuckoo Search (GQCS), Hybrid Cuckoo Search and Differential Evolution (HCSDE), hybrid K-means and Improved Cuckoo Search (KICS), and Quantum Chaotic Cuckoo Search (QCCS) are obtained from the paper [81]. Finally, the performance of the mentioned algorithms is compared in terms of two parameters (SICD and ER) as reported in Table (10). For the sake of clarity, the best results among all clustering methods are illustrated in bold type. It should be noted that the abbreviation "N/A" denotes that no information is available.

**Table 10.** Comparison of the ChOAGNDA and other clustering algorithms from the past literature in terms of SICD and ER values on the clustering datasets.

| Dataset | Method | Year | SICD | | | | ER |
|---|---|---|---|---|---|---|---|
| | | | Best | Mean | Worst | STD | Mean |
| Iris | GSA | 2011 | 96.6970 | 96.7210 | 96.7640 | 0.0123 | 10.04% |
| | GSA-KM | 2012 | 96.6790 | 96.6890 | 96.7050 | 0.0076 | N/A |
| | IKH | 2016 | 96.6555 | 96.6555 | 96.6555 | 0.000009 | 9.78% |
| | KICS | 2017 | 96.7349 | 96.9525 | 97.1901 | 0.1443 | 10.66% |
| | ICMPKHM | 2018 | 96.6123 | 96.6223 | 96.6370 | 0.0105 | 9.76% |
| | SOS | 2019 | 96.6555 | 96.6555 | 96.6555 | 2.82e-14 | N/A |
| | QALO-K | 2020 | 96.6600 | 96.7300 | 97.3500 | N/A | N/A |
| | H-HHO | 2021 | 97.2352 | 98.2954 | 122.365 | 4.3658 | 20.86% |
| | Proposed Method | Present | 96.5400 | **96.5403** | 96.5404 | 0.0001 | **9.10%** |
| Wine | GSA | 2011 | 16315.356 | 16376.619 | 16425.584 | 31.3412 | 29.15% |
| | GSA-KM | 2012 | 16294.250 | 16294.310 | 16294.640 | 0.0406 | N/A |
| | IKH | 2016 | 16292.210 | 16294.300 | 16292.840 | 0.7067 | 28.90% |
| | KICS | 2017 | 16298.627 | 16341.463 | 16437.384 | 37.7510 | 28.98% |
| | ICMPKHM | 2018 | 16292.120 | 16293.180 | 16293.580 | 0.4800 | 28.22% |
| | SOS | 2019 | 16292.184 | 16293.052 | 16294.170 | 0.8185 | N/A |
| | QALO-K | 2020 | 16357.920 | 16453.160 | 16518.750 | N/A | N/A |
| | H-HHO | 2021 | 16121.571 | **16291.219** | 18457.545 | 19.5451 | 33.56% |
| | Proposed Method | Present | 16292.184 | 16292.184 | 16292.6127 | 0.0008 | **26.51%** |
| Cancer | GSA | 2011 | 2967.963 | 2973.583 | 2990.834 | 8.1731 | 3.74% |
| | GSA-KM | 2012 | 2965.140 | 2965.210 | 2965.300 | 0.0670 | N/A |
| | IKH | 2016 | 2964.387 | 2964.389 | 2964.393 | 0.0012 | 3.69% |
| | KICS | 2017 | 2967.216 | 2973.387 | 2982.068 | 4.1859 | 3.80% |
| | ICMPKHM | 2018 | 2965.110 | 3024.790 | 3147.080 | 0.3810 | 3.68% |
| | SOS | 2019 | 2964.387 | 2964.387 | 2964.387 | 3.312e12 | N/A |
| | QALO-K | 2020 | 2964.390 | 2977.930 | 2986.960 | N/A | N/A |
| | H-HHO | 2021 | 2964.386 | 2987.256 | 2998.654 | 19.2654 | 39.11% |
| | Proposed Method | Present | 2964.386 | **2964.3862** | 2964.3870 | 0.0002 | **3.51%** |
| CMC | GSA | 2011 | 5698.156 | 5599.845 | 5702.092 | 1.7244 | 55.67% |
| | GSA-KM | 2012 | 5697.030 | 5697.360 | 5697.870 | 0.2717 | N/A |
| | IKH | 2016 | 5693.720 | 5693.735 | 5693.779 | 0.0079 | 55.90% |
| | KICS | 2017 | 5537.534 | 5540.652 | 5542.182 | 1.6126 | 57.64% |
| | ICMPKHM | 2018 | 5692.110 | 5695.130 | 5697.690 | 0.8120 | 54.41% |
| | SOS | 2019 | 5693.724 | 5693.725 | 5693.728 | 0.0020 | N/A |
| | QALO-K | 2020 | 5542.180 | 5545.050 | 5543.440 | N/A | N/A |
| | H-HHO | 2021 | 5532.188 | 5541.297 | 5554.269 | 1.2954 | 54.10% |
| | Proposed Method | Present | 5532.015 | **5532.184** | 5532.189 | 0.0007 | **49.33** |
| Blood | CS | 2014 | 407942.005 | 408838.235 | 410019.615 | 536.1597 | 34.89% |
| | GQCS | 2014 | 407714.446 | 407716.715 | 407724.582 | 2.2487 | 34.89% |
| | HCSDE | 2015 | 408228.567 | 411305.327 | 416367.360 | 2307.6760 | 34.89% |
| | KICS | 2017 | 407722.178 | 407807.439 | 407986.313 | 71.2328 | 34.89% |
| | QCCS | 2018 | 407714.231 | 407714.231 | 407714.233 | 0.0005 | 34.89% |
| | Proposed Method | Present | 407714.218 | **407714.210** | 407714.280 | 0.0001 | **34.81%** |

Table (10) summarizes the statistical results of our proposed method and previous studies on SICD and ER values obtained from the Iris, Wine, Cancer, CMC, and Blood datasets. From Table (10), it can be observed that the ChOAGNDA significantly outperforms other existing algorithms in terms of average SICD and ER results on the Iris dataset. By inspecting the results in Table (10), it is found that the H-HHO provides the minimum value of SICD compared with other algorithms for the Wine dataset. However, ChOAGNDA has obtained the best value of ER among all competing approaches. For the Cancer dataset, the results confirm the superior performance of ChOAGNDA compared with other algorithms according to the SICD and ER values. Moreover, it is obvious that our proposed method performs much better results than the others in terms of SICD and ER values on the CMC dataset. Lastly, the results of Table (10) show the significant performance of ChOAGNDA among all the considered algorithms regarding the SICD value on the Blood dataset. In terms of the ER results, ChOAGNDA has provided a better value for the Blood dataset.

## 5.5. Statistical Performance Evaluation

The results of subsection 5.4.2 show that the proposed method outperforms other clustering algorithms in terms of SICD and ER results on almost all datasets. To prove the validity of this claim, two statistical techniques (namely: Friedman and Post-Hoc) [82] have been carried out on the mean values of SICD and ER results from 50 independent runs. In particular, non-parametric statistical experiments are performed to determine whether any notable difference exists among the results of various clustering approaches. The Friedman test evaluates the performance of different algorithms according to the principle of the null hypothesis ($H_0$), where the rejection of $H_0$ indicates that the algorithm significantly outperforms other existing algorithms. In this study, different algorithms are ranked based on their SICD and ER results obtained from eight clustering datasets (see Tables (8) and (9)). The approach with the best performance received the lowest rank, while the approach with the worst performance received the highest rank. Equation (29) calculates the Average Rank (AR) of each clustering method according to the Friedman's statistic, and the results are summarized in Tables (11) and (12).

$$AR = \frac{\text{The sum of the total algorithm's ranks for each dataset}}{\text{Total number of datasets}} \quad (29)$$

**Table 11.** Statistical ranking based on the mean values of SICD performances.

| Algorithm | AR |
|---|---|
| K-Means | 8.125 |
| GA | 6.25 |
| PSO | 4.25 |
| MVO | 6.625 |
| GWO | 4.375 |
| ABC | 6 |
| ACO | 5.625 |
| WOA | 2.625 |
| **ChOAGNDA** | **1** |

**Table 12.** Statistical ranking based on the mean values of ER performances.

| Algorithm | AR |
|---|---|
| K-Means | 7.875 |
| GA | 4.5 |
| PSO | 4 |
| MVO | 6.125 |
| GWO | 5.375 |
| ABC | 5.25 |
| ACO | 4.875 |
| WOA | 3.25 |
| **ChOAGNDA** | **1** |

Figures (11) and (12) compare the AR values of each considered algorithm on all datasets based on their SICD and ER results, respectively. It can be observed that the proposed ChOAGNDA is ranked as the first-best algorithm among all existing methods.

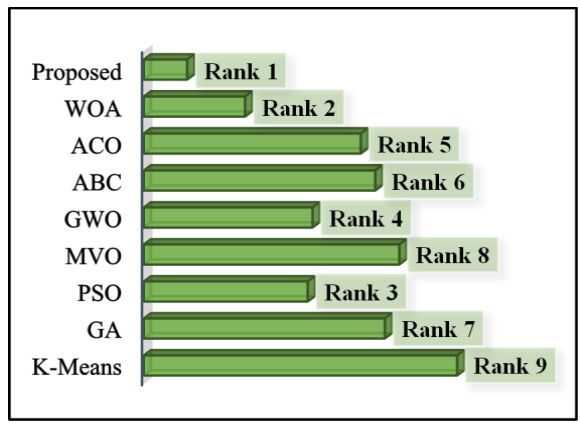
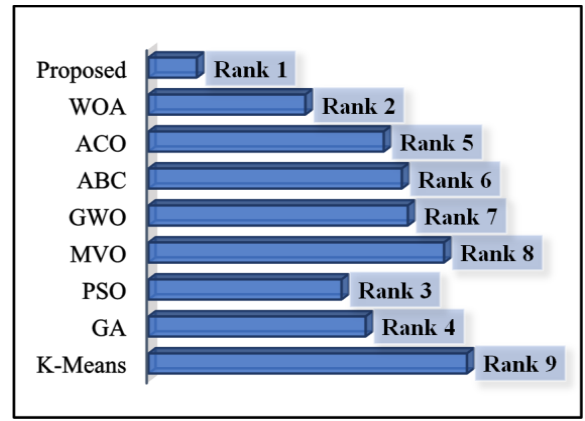

**Figure 11.** Average ranking of the clustering algorithms according to the SICD results using Friedman test.

**Figure 12.** Average ranking of the clustering algorithms according to the ER results using Friedman test.

In the following, the statistical validations are conducted to compare the performance of the ChOAGNDA with other clustering algorithms. It should be noted that the level of significance in our case is $\propto=0.05$. The calculated Friedman statistic on the SICD results is 39.43, and the corresponding p-value is 0.000004084. Consequently, the null hypothesis is rejected, meaning that the proposed method performs significantly better than the other approaches. All existing algorithms are further analyzed using the Post-Hoc test to confirm the statistical superiority of the proposed method. Table (13) reports the findings achieved by the Post-Hoc procedure over the SICD results, while ChOAGNDA is considered as the control algorithm. The results clearly indicate that the null hypothesis is rejected in the majority of pairwise comparisons.

**Table 13.** Statistical results of Post-Hoc test according to the SICD values ($\propto=0.05$).

| Algorithm | Z-value | P-value | Null Hypothesis ($H_0$) |
|---|---|---|---|
| ChOAGNDA vs K-Means | 5.112077 | 0.000000318 | Rejected |
| ChOAGNDA vs GA | 3.742771 | 0.000182002 | Rejected |
| ChOAGNDA vs PSO | 2.282177 | 0.002247887 | Rejected |
| ChOAGNDA vs MVO | 4.016632 | 0.000059035 | Rejected |
| ChOAGNDA vs GWO | 2.373464 | 0.017622091 | Rejected |
| ChOAGNDA vs ABC | 3.560197 | 0.000370577 | Rejected |
| ChOAGNDA vs ACO | 3.286335 | 0.001500101 | Rejected |
| ChOAGNDA vs WOA | 2.276335 | 0.273321730 | Not Rejected |

On the other hand, the calculated Friedman statistic on the ER results is 36.09, and the corresponding p-value is 0.0000689. Hence, the null hypothesis is rejected at the 0.05 level of significance. Afterward, the Post-Hoc test is applied over the ER results to confirm whether the suggested method is significant or not. Table (14) represents the experimental results of the Post-Hoc procedure. From Table (14), we can say that the ChOAGNDA is statistically considerable in comparison to the other meta-heuristic algorithms. By reconsidering all the reported results, it can be concluded that the proposed method outperforms other clustering techniques in terms of the SICD results, ER values, convergence rates, and statistical performances.

**Table 14.** Statistical results of Post-Hoc test according to the ER values ($\alpha=0.05$).

| Algorithm | Z-value | P-value | Null Hypothesis ($H_0$) |
|---|---|---|---|
| ChOAGNDA vs K-Means | 5.340295 | 0.000000092 | Rejected |
| ChOAGNDA vs GA | 2.875543 | 0.004033327 | Rejected |
| ChOAGNDA vs PSO | 2.327821 | 0.001992162 | Rejected |
| ChOAGNDA vs MVO | 4.016632 | 0.000059035 | Rejected |
| ChOAGNDA vs GWO | 3.331979 | 0.000862308 | Rejected |
| ChOAGNDA vs ABC | 3.377622 | 0.000731154 | Rejected |
| ChOAGNDA vs ACO | 2.966831 | 0.003886803 | Rejected |
| ChOAGNDA vs WOA | 1.643168 | 0.100348221 | Not Rejected |

## 6. Conclusion

Recently, the application of meta-heuristic algorithms has grown substantially due to the limitations of classical techniques in solving clustering problems. Classical algorithms have some disadvantages associated with getting stuck in local optima or suffering from slow convergence speed in high-dimensional datasets. To overcome these deficiencies, this paper suggested a new technique for solving clustering problems based on the chimp optimization algorithm (ChOA), Generalized Normal Distribution Algorithm (GNDA), and Opposition-based Learning (OBL) technique, called ChOAGNDA. The performance of the proposed algorithm was evaluated on eight benchmark datasets, and the experimental results were compared against several well-known clustering algorithms. The effectiveness of the approaches was further investigated according to the three measurements: SICD result, ER value, and convergence rate. Lastly, some statistical tests have been conducted to verify that the obtained dissimilarities between the ChOAGNDA and other algorithms are significantly meaningful. Experimental results proved the superior performance of the proposed clustering technique on the majority of evaluation criteria.

The main contributions of this study provide several enhancements for solving data clustering problems. First of all, ChOAGNDA contains a limited number of parameters to adjust. Secondly, independent groups of chimps in ChOA assist ChOAGNDA algorithm in dealing with complex clustering tasks. Thirdly, selecting a proper chaotic map can resolve two significant issues: slow convergence rate and trap in local minima. Moreover, not only the dynamic coefficient of the f vector guarantees to reach an appropriate balance between the local and global search but also the OBL technique allows ChOAGNDA for a stable balance between exploration and exploitation phases to avoid local optima. Finally, the proposed algorithm outperformed the existing clustering methods in terms of SICD and ER results. This study opens up several research directions for future work in a variety of scientific fields and real-world problems. Accordingly, the proposed clustering algorithm can be widely used for many applications, such as: (i) fraud detection: identification of fraud risk through text mining, (ii) real-life examples: market and customer segmentation, (iii) biomedical fields: medical-image analysis, (iv) engineering sciences: recommendation systems, (v) health informatics: biomedical data analysis and identification of cancer cells, and (vi) industrial domains: identification of heavy metal pollution sources. For the future direction, various applications of ChOAGNDA algorithm are highly recommended. We believe that the proposed hybrid approach will be beneficial for solving some complex real-world problems.

**CRediT authorship contribution statement**

**Sayed Pedram Haeri Boroujeni**: Methodology, Formal analysis, Software, Validation, Writing. **Elnaz Pashaei**: Conceptualization, Supervision

**Declaration of competing interest**

The authors declare that this research did not receive any specific grant from funding agencies in the public, commercial, or not-for-profit sectors.

**Acknowledgment**

The authors would like to thank anonymous reviewers for their time, valuable comments, constructive criticism, and suggestions, which greatly improved the paper.